%% file: icml2023.tex
\documentclass{article} 

\usepackage[accepted]{icml2023}
\usepackage{times}

\input{math_commands.tex}

\usepackage[hidelinks]{hyperref}
\usepackage{url}

\usepackage{adjustbox}
\usepackage{wrapfig}
\usepackage{microtype}
\usepackage{graphicx}
\usepackage{subfigure}
\usepackage{booktabs} 

\usepackage{amsmath}
\usepackage{amssymb}
\usepackage{mathtools}
\usepackage{amsthm}
\usepackage{multirow}
\usepackage[ruled, vlined, linesnumbered, algo2e]{algorithm2e} 

\usepackage{enumitem}

\SetKwInOut{Parameter}{Parameters}

\usepackage{caption}

\usepackage[capitalize,noabbrev]{cleveref}

\begin{document}
\twocolumn[
\icmltitle{Can we achieve robustness from data alone?}


\begin{icmlauthorlist}
\icmlauthor{Nikolaos Tsilivis}{yyy}
\icmlauthor{Jingtong Su}{yyy}
\icmlauthor{Julia Kempe}{yyy}
\end{icmlauthorlist}

\icmlaffiliation{yyy}{Center for Data Science, New York University}

\icmlcorrespondingauthor{Nikolaos Tsilivis}{nt2231@nyu.edu}

\icmlkeywords{Machine Learning, ICML}

\vskip 0.3in
]
\printAffiliationsAndNotice{}

\begin{abstract}

We introduce a meta-learning algorithm for adversarially robust classification. The proposed method tries to be as model agnostic as possible and optimizes a dataset prior to its deployment in a machine learning system, aiming to effectively erase its non-robust features. Once the dataset has been created, in principle no specialized algorithm (besides standard gradient descent) is needed to train a robust model. We formulate the data optimization procedure as a bi-level optimization problem on kernel regression, with a class of kernels that describe infinitely wide neural nets (Neural Tangent Kernels). We present extensive experiments on standard computer vision benchmarks using a variety of different models, demonstrating the effectiveness of our method, while also pointing out its current shortcomings. In parallel, we revisit prior work that also focused on the problem of data optimization for robust classification \citep{Ily+19}, and show that being robust to adversarial attacks after standard (gradient descent) training on a suitable dataset is more challenging than previously thought.

\end{abstract}

\section{Introduction}
\label{sec:intro}

The discovery of the adversarial vulnerability of neural nets \citep{Sze+14,GSS15,Pap+17,CaWa17} - their brittleness when exposed to imperceptible perturbations in the data - 
has shifted the focus of the machine learning community from standard gradient techniques to more complex training algorithms that are rooted in robust optimization. In principle, if $\mathcal{P}$ denotes a data distribution and $\Delta$ is a set of allowed perturbations of the input space, we would like to solve the following problem \citep{Mad+18}:
\begin{equation}\label{eq:rob_opt}
    \inf_{\theta} \mathbb{E}_{(x, y) \sim \mathcal{P}} \sup_{\delta \in \Delta} \mathcal{L} (f (x + \delta; \theta, \mathcal{D}_{\mathrm{train}}), y),
\end{equation}
where $f$ is a model parameterized by $\theta$ (e.g. a neural network), $\mathcal{D}_{\mathrm{train}}$ denotes a finite dataset used for training, and $\mathcal{L}$ is a loss function used for classification.

Since solving this problem is generally intractable, it is common to employ an iterative algorithm that interchangeably performs steps of gradient ascent/descent, a procedure called adversarial training \citep{GSS15,Mad+18}. Here, training data is augmented on the fly through perturbations coming from the very model it is training. While adversarial training in its many variants (e.g. \citep{,Zha+19,WRK20}) has been successful and currently constitutes the only defense that consistently withstands adversarial attacks, it is computationally expensive and the produced models still fall short in absolute accuracy \citep{Cro+20}.

However, once we focus on non-parametric models $f$ (such as kernel ridge regression), we can pose a more ``direct'' problem
\begin{equation}\label{eq:data_framework}
    \inf_{\mathcal{D}_{\mathrm{train}}} \mathbb{E}_{(x, y) \sim \mathcal{P}} \sup_{\delta \in \Delta} \mathcal{L} (f (x + \delta; \mathcal{D}_{\mathrm{train}}), y),
\end{equation}
where instead of optimizing the model parameters, we optimize the {\em training data}.
The above formulation has the benefit of directly optimizing  the quantity of interest, that is the robust loss at the end of ``training''/deployment. Additionally, since the outcome of this optimization is a dataset, it can be deployed with any other model, and, given favorable transfer properties, might yield good performance even outside the scope 
it was optimized for, without the need for costly adversarial training of the new model. This latter hope is not unfounded, since adversarial examples themselves have been shown to be rather universal and transferable across models \citep{Pap+17,Dez+17}. 

Datasets are more modular than models in terms of compatibility with different frameworks and codebases; and \textit{robust} datasets could accelerate research in trustworthy machine learning. Recent work has explored the role of data in many machine learning areas. \citet{Wang18} have initiated a growing body of work on dataset distillation. Data pruning techniques \citep{Diet21} have been successfully used for finding sparse networks \citep{paul2022lottery} or improving neural scaling laws \citep{sorscher2022beyond}, making our exploration into data-induced robustness even more timely.


\begin{figure*}
    \centering
    \includegraphics[scale=0.2]{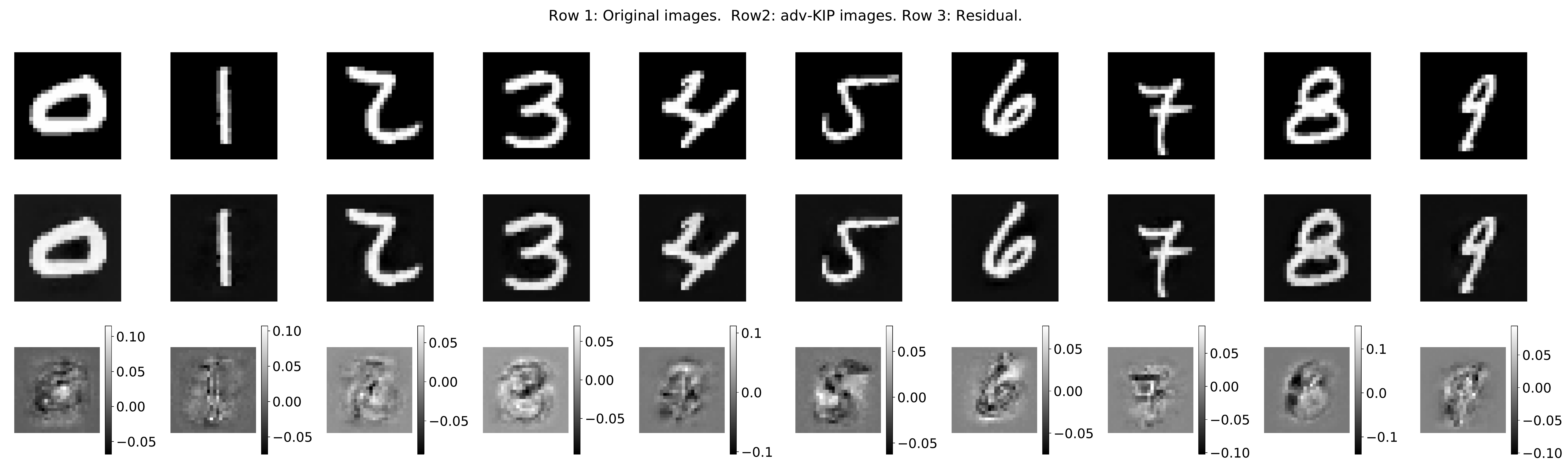}
    \caption{Samples from MNIST - one per class, before (top row) and after (middle row) optimization with \texttt{adv-KIP}. The bottom row shows the residual between the images of that column. Performance of NTK kernel regression on original data: 98.46\% clean test accuracy, 0.07\% robust test accuracy. On optimized data: 97.67\% clean test accuracy, 42.94\% robust test accuracy. \texttt{adv-KIP} hyperparameters: $\epsilon=0.3$, 10 PGD steps, $\alpha=0.1$, fully connected ReLU NTK with 3 layers, dataset size is 40,000.}
    \label{fig:ad_figure}
\end{figure*}

In this work, we initiate such a data-based approach to robustness in a principled way. We propose a gradient-based approach for solving the optimization problem in Eq.~(\ref{eq:data_framework}), adopting kernel regression; focusing on a particular class of kernel functions, \textit{Neural Tangent Kernels} (NTKs).
Kernel regression with NTKs is known to describe the training process of infinitely wide networks \citep{JHG18,Lee+19,Aro+19b}, providing a natural choice for the optimization of datasets for eventual deployment in neural networks. Prior work on dataset distillation (the process of distilling information from one dataset to a smaller, optimized, one) has found that meta-optimization using NTKs produces highly transferable datasets suitable for neural network training \citep{Ngu+21a,Ngu+21b}. Their algorithm, \textit{Kernel Inducing Point} (KIP), serves as an inspiration for this work; thus we give the name \texttt{adv-KIP} to our novel meta-learning algorithm for iteratively solving Eq.~(\ref{eq:data_framework}).

We perform a wide range of experiments on standard image recognition datasets (MNIST and CIFAR-10) using \texttt{adv-KIP} with a variety of different kernels and perturbation sets $\Delta$, and discuss important design choices (dataset size, meta-optimizer, loss choices etc). We establish that the meta-learning algorithm converges to training datasets which look similar to the original ones (see Fig.~\ref{fig:ad_figure}), when used as a training set guarantee similar performance on an (unperturbed) test set, but additionally, and importantly, are also able to produce models capable of defending against gradient-based attacks.

However, we find that the models struggle to withstand non-gradient based, or adaptive, attacks. Interestingly, we observe that both neural networks and kernels that are trained on our optimized datasets have vanishing gradients with respect to the input, thus causing gradient based attacks to fail. This phenomenon, known in the literature under the umbrella term of \textit{obfuscated gradient} \citep{Ath+18}, has been observed in the past, but has been largely connected to (failed) defensive interventions on models and architectures (stochastic/discrete layers etc.~). Our findings show that it can also be observed in more ``benign" settings, and can also be a property of the data alone.

Armed with these insights, we revisit seminal work of \citet{Ily+19} that put forward a dichotomy between \textit{robust and non-robust} features \citep{Tsi+19,allen2022feature,TsilivisKempe22}
in the data. To bolster their argument, \citet{Ily+19} construct a dataset (using an already adversarially trained model) thought to contain solely robust features and demonstrate that a freshly initialized neural network (standardly) trained on this dataset is able to achieve non-trivial performance against gradient based attacks. Remarkably, upon re-examination, we find that models trained with these optimized datasets also suffer from the vanishing gradient problem and fail to withstand adaptive attacks, thus shattering a long standing claim in the community. Collectively, our findings establish that adversarial robustness from data alone is far more challenging than previously thought. We believe that our insights will help design and faster debug other data-based approaches in the future.


In summary, our contributions are two-fold:
\begin{itemize}[topsep=0pt]
 \setlength\itemsep{-3pt}
    \item We devise a principled data based meta-learning algorithm, \texttt{adv-KIP}, tackling the bi-level optimization for adversarially robust classification in a novel way, with potentially  wider applicability. We perform experiments on standard computer vision benchmarks and analyze the properties of models (kernels and neural networks) fit on the optimized datasets demonstrating that they enjoy remarkable robustness against gradient-based attacks.

    \item We show that robustness from data alone is more challenging than previously assumed. Standard models trained with gradient descent on datasets from \texttt{adv-KIP} suffer from the ``obfuscated gradient" phenomenon, without explicitly adding non-smooth elements to the computational pipeline. We find that this is also the case for models trained with datasets presumably containing solely robust features \citep{Ily+19}. That is, all these models fail to withstand non-gradient based adversarial attacks, once the size of the perturbation is large enough. We examine and discuss the properties of all the models, in terms of confidence and calibration and draw lessons for future dataset design.
\end{itemize}


\section{Preliminaries}\label{sec:prelim}

 \textbf{Adversarial Training.} 
Eq. \ref{eq:rob_opt} establishes the min-max underpinning for the construction of adversarially robust classifiers \citep{Mad+18}. The most common way to approximate the solution of this optimization problem for a neural network $f$ is to first generate adversarial examples by running multiple steps of projected gradient descent (PGD) \citep{KGB17,Mad+18}. When the set of allowed perturbations $\Delta$ is ${\mathcal{B}_{\mathbf{x}}^\epsilon}$ - the $\ell_\infty$ ball of radius $\epsilon$ and center $\mathbf{x}$ - the iterative $N$-step approximation is given by
\begin{equation}\label{eq:pgd}
    \mathbf{x}^{k+1} = \Pi_{\mathcal{B}_{\mathbf{x}^0}^\epsilon} \left( \mathbf{x}^k + \alpha \cdot \mathrm{sign} (\nabla_{\mathbf{x}^k} \mathcal{L}(f(\mathbf{x}^k), y) \right),
\end{equation}
where $\mathbf{x}^0 = \mathbf{x}$ is the original example, $\alpha$ is a learning step, $\tilde{\mathbf{x}} = \mathbf{x}^N$ is the final adversarial example, and  $\Pi$ is the projection on the valid constraint set of the data.
During adversarial training we alternate steps of generating adversarial examples (using $f$ from the current network) and training on this data instead of the original one. Several variations of this approach have been proposed in the literature (e.g. \cite{Zha+19,Sha+19,Wong+20}), modifying either the attack used for data generation (inner loop in Eq.~(\ref{eq:rob_opt})) or the loss in the outer loop.

\textbf{Kernel Regression, NTK and KIP.}
Kernel regression is a fundamental non-linear regression method.
Given a dataset $(\mathcal{X}, \mathcal{Y})$
, where $\mathcal{X} \in \mathbb{R}^{n \times d}$ and $\mathcal{Y} \in \mathbb{R}^{n \times k}$ (e.g., a set of one-hot vectors), kernel regression computes an estimate
\begin{equation}\label{eq:kernel_reg}
    \hat{f}(\mathbf{x}) = K(\mathbf{x}, \mathcal{X})^\top K(\mathcal{X}, \mathcal{X})^{-1} \mathcal{Y},
\end{equation}
where $K(\mathbf{x}, \mathcal{X}) = [k(\mathbf{x}, \mathbf{x}_1), \ldots, k(\mathbf{x}, \mathbf{x}_n)]^\top \in \mathbb{R}^{n}$, $K(\mathcal{X}, \mathcal{X})_{ij} = k(\mathbf{x}_i, \mathbf{x}_j)$ and $k$ is a kernel function that measures similarity between points in $\mathbb{R}^d$.

Recent work in deep learning theory has established a profound connection between kernel regression and the infinite width, low learning rate limit of deep neural networks \citep{JHG18,Lee+19,Aro+19b}. In particular, it can be shown that the evolution of such suitably initialized infinitely wide neural networks admits a closed form solution as in Eq.~(\ref{eq:kernel_reg}), with a network-dependent kernel function $k$. Focusing on a scalar neural net $f$ for ease of notation, it is given by:
\begin{equation}
    k(\mathbf{x}_i, \mathbf{x}_j) = \nabla_{\theta} f(\mathbf{x}_i; \theta)^\top \nabla_{\theta} f(\mathbf{x}_j; \theta),
\end{equation}
where $\theta$ are the parameters of the network. This expression becomes constant (in time) in the infinite width limit.

Many fruitful applications of NTK theory to machine learning practice have already benefited from the equivalence between kernels and neural networks \citep{Tan+20,Chen+21,Ngu+21b}. Our work builds on a recently proposed dataset distillation \citep{Wan+18} algorithm called \textit{Kernel Inducing Points} (KIP) \citep{Ngu+21a,Ngu+21b}. These works introduce a meta-learning algorithm for data distillation from an original training set $\mathcal{D}$, to an optimized {\em source} set $(\mathcal{X}_S,\mathcal{Y}_S)$ of \textit{reduced} size but of similar generalization properties. The closed form of Eq.~(\ref{eq:kernel_reg}) allows to express this objective via a loss function on a {\em target} data set $(\mathcal{X}_T,\mathcal{Y}_T)$ as:
\begin{equation}\label{eq:kip_loss}
    \mathcal{L}_{\mathrm{KIP}} (\mathcal{X}_S,  \mathcal{Y}_S) = \| \mathcal{Y}_T - K(\mathcal{X}_T, \mathcal{X}_S)^\top K(\mathcal{X}_S, \mathcal{X}_S)^{-1} \mathcal{Y}_S \|_2.
\end{equation}
The error of Eq.~(\ref{eq:kip_loss}) can be minimized via gradient descent on $\mathcal{X}_S$ (and optionally $\mathcal{Y}_S$). Starting with a smaller subset of $\mathcal{D}$, sampling a target dataset from $\mathcal{D}$ to simulate test points, and backpropagating the gradients of the error with respect to the data allows to progressively find better and better synthetic data.
Importantly, leveraging the NTK for kernel regression renders the datasets suitable for deployment on actual neural nets as well.

 \textbf{Prior work on dataset optimization.} 
To the best of our knowledge, the idea of trying to obtain robust classifiers through data or representation optimization is rather unexplored. \citet{Gar+18} design a spectral method to extract robust embeddings from a dataset. \citet{Awa+21} formulate an adversarially robust formulation of PCA, to extract provably robust representations. \citet{Ily+19} construct a robust dataset by traversing the representation layer of a previously trained robust classifier, which presumably contains solely robust features. Yet, all of these methods achieve substantially lower robust accuracy compared to adversarial training.

\section{adv-KIP Algorithm}
\label{approach}

In this section, we introduce our meta-learning approach, \texttt{adv-KIP}, for adversarially robust classification. It optimizes a dataset prior to its deployment so that a model fit on it will have  predictions that are robust against worst case perturbations of a testing input. That is, our algorithm inputs a dataset, outputs a dataset and carries out optimization (i.e. gradient updates) on the data level. Similarly to the KIP method \citep{Ngu+21a,Ngu+21b} outlined in Section \ref{sec:prelim}, our method works with kernel machines, and especially with NTKs. However, the goal here is slightly different from the KIP objective; instead of deriving a dataset of reduced size, we aim to create one that induces better robustness properties \textit{on the original unmodified test set}. For further connections with previous work in adversarially robust classification, see App.~\ref{app:previouswork}.

We modify the meta-objective of Eq.~(\ref{eq:kip_loss}), and instead of optimizing the data $(\mathcal{X}_S, \mathcal{Y}_S)$ with respect to the ``clean" loss of Eq.~(\ref{eq:kip_loss}), we minimize
\begin{equation}\label{eq:adv_kip_loss}
    \Tilde{\mathcal{L}} (\mathcal{X}_S,  \mathcal{Y}_S) = 
     \| \mathcal{Y}_T - K(\tilde{\mathcal{X}}_T, \mathcal{X}_S)^\top K(\mathcal{X}_S, \mathcal{X}_S)^{-1} \mathcal{Y}_S \|_2,
\end{equation}
where, in a slight abuse of notation, $\tilde{\mathcal{X}}_T = \mathcal{X}_T + \tilde{\delta}$, and
\begin{equation}\label{eq:adv_kip_max}
    \tilde{\delta} = \arg \max_{\delta \in \Delta} \mathcal{L}(K(\tilde{\mathcal{X}}_T, \mathcal{X}_S)^\top K(\mathcal{X}_S, \mathcal{X}_S)^{-1} \mathcal{Y}_S, \mathcal{Y}_T).
\end{equation}

In other words, we optimize the data $(\mathcal{X}_S, \mathcal{Y}_S)$ to achieve the best possible performance of kernel regression on the target data set \textit{after the adversarial attack}.
Notice that this approach solves the problem outlined in Eq.~(\ref{eq:data_framework}), by adding an inner maximization problem to the KIP framework. Solving this optimization now requires an inner loop that tackles the maximization in Eq.~(\ref{eq:adv_kip_max}). As is common, we employ first order gradient methods to solve this, namely projected gradient ascent (on the loss of the kernel machine).  Algorithm \ref{alg:adv-kip} describes our generic robust training data set distillation framework for the case of an $\ell_\infty$ threat model, though it is easily modified to any constraint set $\Delta$.

\begin{algorithm2e}
    \caption{\texttt{adv-KIP} for $\ell_\infty$ adversarial attacks}
    \KwIn{A training dataset $\mathcal{D}_{\mathrm{train}} = \{ \mathcal{X}, \mathcal{Y} \}$.}
    \Parameter{Number of (meta) $\mathrm{epochs}$, number of $\mathrm{pgd\_steps}$, pgd step size $\alpha$, max attack size $\epsilon$, (meta) learning rate $\lambda$.}
    \KwOut{A new dataset $\mathcal{D}_{\mathrm{rob}}$.}
    Sample data $\mathcal{S} = \{\mathcal{X}_S, \mathcal{Y}_S\}$ from $\mathcal{D}_{\mathrm{train}}$\;
    \For{$\mathrm{i} \gets 1$ \KwTo $\mathrm{epochs}$}{
    Sample data $\mathcal{T} = \{\mathcal{X}_T, \mathcal{Y}_T\}$ from $\mathcal{D}_{\mathrm{train}}$\;
        \For{$\mathrm{j} \gets 1$ \KwTo $\mathrm{pgd\_steps}$}{
        $\mathcal{X}_T \gets \mathcal{X}_T + \alpha \cdot \mathrm{sign} (\nabla_{\mathcal{X}_T} \mathcal{L}_{\mathrm{ce}}(K_{\mathcal{X}_T \mathcal{X}_S} K_{\mathcal{X}_S \mathcal{X}_S}^{-1} \mathcal{Y}_S, \mathcal{Y}_T))$\;
        $\mathcal{X}_T \gets \Pi_{\mathcal{B}_\epsilon}(\mathcal{X}_T)$\;
        }
        $\mathcal{X}_S \gets \mathcal{X}_S - \lambda \nabla_{\mathcal{X}_S} \mathcal{L}(K_{\mathcal{X}_T \mathcal{X}_S} K_{\mathcal{X}_S \mathcal{X}_S}^{-1} \mathcal{Y}_S, \mathcal{Y}_T)$\;
        $\mathcal{Y}_S \gets \mathcal{Y}_S - \lambda \nabla_{\mathcal{Y}_S} \mathcal{L}(K_{\mathcal{X}_T \mathcal{X}_S} K_{\mathcal{X}_S \mathcal{X}_S}^{-1} \mathcal{Y}_S, \mathcal{Y}_T)$\;
    }
    $\mathcal{D}_{\mathrm{rob}} \gets (\mathcal{X}_S, \mathcal{Y}_S)$
    \label{alg:adv-kip}
\end{algorithm2e}

\textbf{Algorithmic choices / Hyperparameters:} There are several options to specialize in Algorithm \ref{alg:adv-kip}:

    {\em Outer loss function (lines 7 and 8)}: This can be either the mean squared error (as in Eq.~(\ref{eq:adv_kip_loss})) or the more flexible cross entropy loss. Both quantify the discrepancy between the prediction of the kernel machine on the perturbed data and the ground truth labels. Alternatively, a loss that balances clean and robust accuracy (see e.g. \cite{Zha+19}) can be used.

    {\em Inner loss function (line 5)}: Instead of using the cross entropy loss for the generation of adversarial examples, we can also adopt previously proposed methods, like the CW-loss \citep{CaWa17}.
   
    {\em Optimization of labels (line 8)}: This is related to an interesting question raised in prior work: How important are rigid one-hot labels for robust classification (see e.g. \citep{Pap+16} for an ultimately unsuccessful defense). Algorithm \ref{alg:adv-kip} as presented here optimizes the labels, by backpropagating the loss gradients. It has the flexibility, however, to omit this part (by omitting line 8).

    {\em Dataset Size $|\mathcal{X}_S|$}: While the goal of the algorithm is to ultimately produce a robust dataset, a reduced dataset size could be a useful by-product, by initializing with $|\mathcal{X}_S| < |\mathcal{X}|$ samples.
    
    {\em Meta-Optimizer (lines 7 and 8)}: Algorithm \ref{alg:adv-kip} presents the simplest possible version of \texttt{adv-KIP}; but the (meta) optimization of the data $\{\mathcal{X}_S, \mathcal{Y}_S\}$ can also be implemented with other algorithms, such as adaptive gradient methods. Indeed, in the experiments, following the code implementation of KIP \citep{Ngu+21a}, we adopt the Adam optimizer \citep{KiBa15} instead of gradient descent. 
    

\textbf{Algorithmic Framework for Broader Data-Based Problems:} Algorithm \ref{alg:adv-kip} is an approach to a particular min-max problem given in Eq.~(\ref{eq:data_framework}). The advantage of deploying the NTK here is that it affords an analytic surrogate expression for the output of the trained neural networks, which allows to compute gradients with respect to the input dataset. We expect that our framework can be adopted and be helpful in other problems in data-centric machine learning, such as, for instance, average case robustness \citep{HeDi19}, discrimination against minority groups \citep{Hash+18}, few-shot learning (where the inner loop would optimize for accuracy on a small out-of-distribution target set) and possibly others. 

\section{Experiments}\label{sec:experiments}

We evaluate the performance of \texttt{adv-KIP} on standard computer vision benchmarks. Specifically, we perform experiments on MNIST \citep{deng2012mnist} and CIFAR-10 \citep{Krizhevsky09CIFAR} against $\ell_\infty$ and $\ell_2$ adversaries. In all cases, robust performance is measured on the original test dataset. We experiment with kernels that correspond to fully connected and convolutional architectures of relatively small depth, due to the high computational cost of evaluating the NTK expressions. All our experiments are based on expressions available through the Neural Tangents library \citep{Nov+20}, built on top of JAX \citep{Bra+18}. Adversarial attacks are implemented using the CleverHans library \citep{Pap+18} (or slightly modified versions for efficient deployment on kernels). \texttt{FCd} and \texttt{Convd} denote the NTK of a $d$-layer infinitely wide fully-connected ReLU neural network and (fully) convolutional ReLU neural network with an additional fully connected last layer, respectively.

\subsection{Meta-Learning and Kernel results}\label{ssec:results}

\begin{figure}
    \centering
    \includegraphics[scale=0.21]{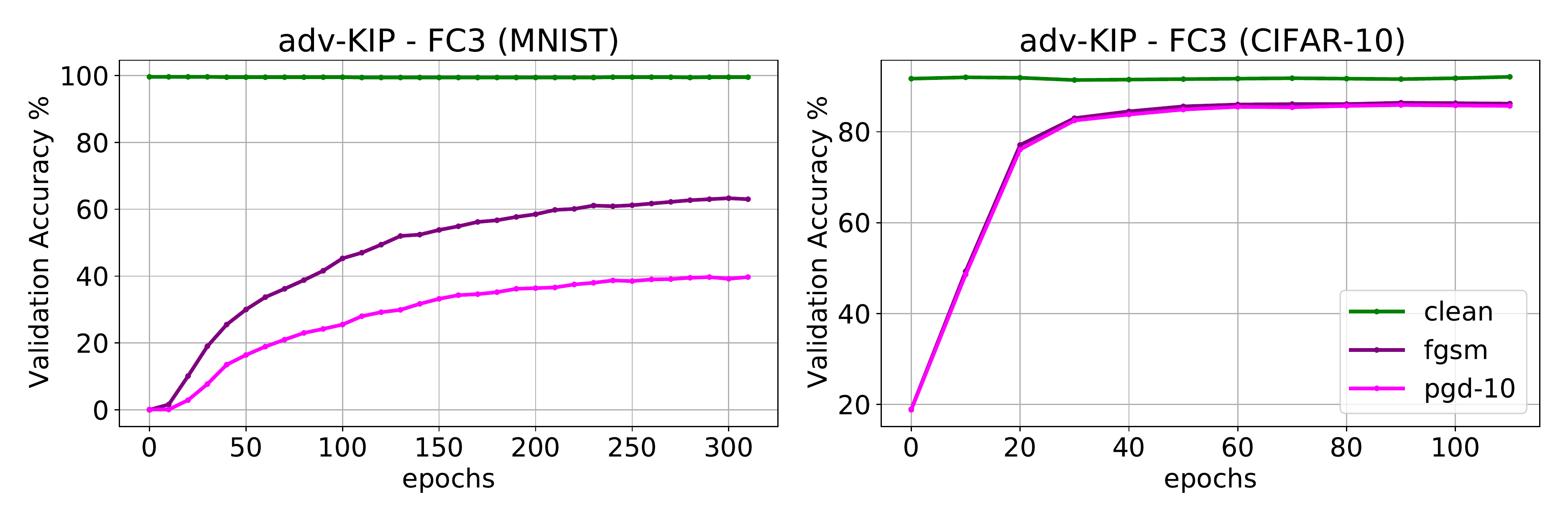}
    \caption{Validation accuracies during meta-optimization with \texttt{adv-KIP}. Setting:
    \texttt{FC3} kernel, $|\mathcal{X}_S|=40K$, $|\mathcal{X}_T|=10K$, $\ell_\infty$ adversary, (left) MNIST $\epsilon=0.3$, $\alpha=0.1$, 10 PGD steps, (right) CIFAR-10 $\epsilon=8/255$, $\alpha=2/255$, 10 PGD steps.}
    \label{fig:kernel_convergence}
\end{figure}

We implement Algorithm \ref{alg:adv-kip} with the Adam optimizer (with fixed learning rate equal to $1e$-3) for a few hundred epochs with multi-step PGD attacks, using standard values of perturbation budget $\epsilon, \alpha$. The datasets are initialized as balanced, i.e. they contain the same number of samples from each class. The performance of the algorithm is evaluated every few epochs on a hold out validation set (part of the original training dataset), and we terminate if validation robust accuracy ceases to increase.

For both MNIST and CIFAR-10, we observe that \texttt{adv-KIP} is able to converge in a few epochs of meta-training (see Fig.~\ref{fig:kernel_convergence}). Validation accuracy against FGSM and PGD attacks increases with the number of outer loop steps, essentially without compromising performance on clean data. Note that at the start of the optimization (corresponding to the original data) the robustness of the kernel machine is very low, as expected from studies on neural nets. 

Table \ref{tab:merge-kernel} summarizes the test results on kernel machines. All kernel classifiers are fit with the \texttt{adv-KIP}-produced dataset meta-optimized with the same kernel. We see that just 5,000 images on MNIST suffice to yield a \texttt{Conv3} classifier scoring $96.31\%$ test accuracy and $76.62\%$ robust test accuracy (evaluated against strong PGD attacks of same strength as used in the inner-loop optimization with radius $\epsilon = 0.3$). Notice that all the robust benefit comes from the new, optimized images (no change in the model). We also note that the convolutional kernel achieves better performance, despite the fact that we deploy it with a much smaller dataset $\mathcal{X}_S$ of size $5,000$. This indicates that convolutional architectures might be more amenable to meta-optimization than their fully connected counterparts. On CIFAR-10, we see a marked drop in both clean and robust accuracy; note, however, that fully connected kernels are not very competitive classifiers for CIFAR-10 images. To achieve {\em some} level of robustness with these simple architectures gives credence to our approach. To the best of our knowledge, this is a first (empirical) demonstration of some robustness of simple machine learning models, such as kernel methods, to adversarial attacks.

As an ablation study, we measure robust accuracy on datasets produced by the original KIP algorithm \citep{Ngu+21a,Ngu+21b} which is designed for reduction of dataset size, while keeping (clean) accuracy as uncompromised as possible. It could be reasonable to hypothesize that such information compression might possibly lead to an increase of robustness as well, but we find that this is not at all the case. We also experimented with larger KIP datasets of the same cardinality as the ones of Table \ref{tab:merge-kernel}. In all cases, we find that KIP datasets do not provide any robust accuracy, neither against FGSM nor PGD attacks (see Appendix \ref{app:KIP} for more details). This indicates the clear need to adjust the optimization objective to robust performance, as is done in the \texttt{adv-KIP} algorithm.

\begin{table}[h]
    \caption{Test accuracy $\%$ (clean and robust) of kernel machines using the optimized \texttt{adv-KIP} datasets. For MNIST ($\ell_\infty$): $\epsilon=0.3$, $\alpha=0.1$, $10$ PGD steps. For CIFAR-10 ($\ell_\infty$): $\epsilon=8/255$, $\alpha=2/255$, $6$ PGD steps. For CIFAR-10 ($\ell_2$): $\epsilon=1.$, $\alpha=0.2$, $10$ PGD steps. These exact same hyperparameters are used for attacks in the inner loop of the dataset meta-optimization.}
    \label{tab:merge-kernel}\ 
    \centering
    \resizebox{0.9\columnwidth}{!}{
    \begin{tabular}{ccccc}
        \toprule
       & & & \multicolumn{2}{c}{\textbf{Robust}} \\
        \textbf{Dataset} & \textbf{Kernel}, $|\mathcal{X}_s|$ & Clean & FGSM & PGD \\
        \midrule
        \multirow{2}{*}{MNIST ($\ell_\infty$)} & \texttt{Conv3}, 5k & 96.31 & 94.82 & 76.62 \\
        & \texttt{FC7}, 30k & 97.21 & 67.04 & 50.34 \\
        \midrule
        \multirow{2}{*}{CIFAR-10 ($\ell_\infty$)} & \texttt{FC2}, 40k & 59.65 & 20.49 & 20.37  \\
        & \texttt{FC3}, 40k & 60.14 & 36.44 & 36.30 \\
        CIFAR-10 ($\ell_2$) & \texttt{FC3}, 40k & 59.84 & 29.90 & 29.63 \\
        \bottomrule
    \end{tabular}}
\end{table}

\begin{figure}
    \centering
    \includegraphics[scale=0.21]{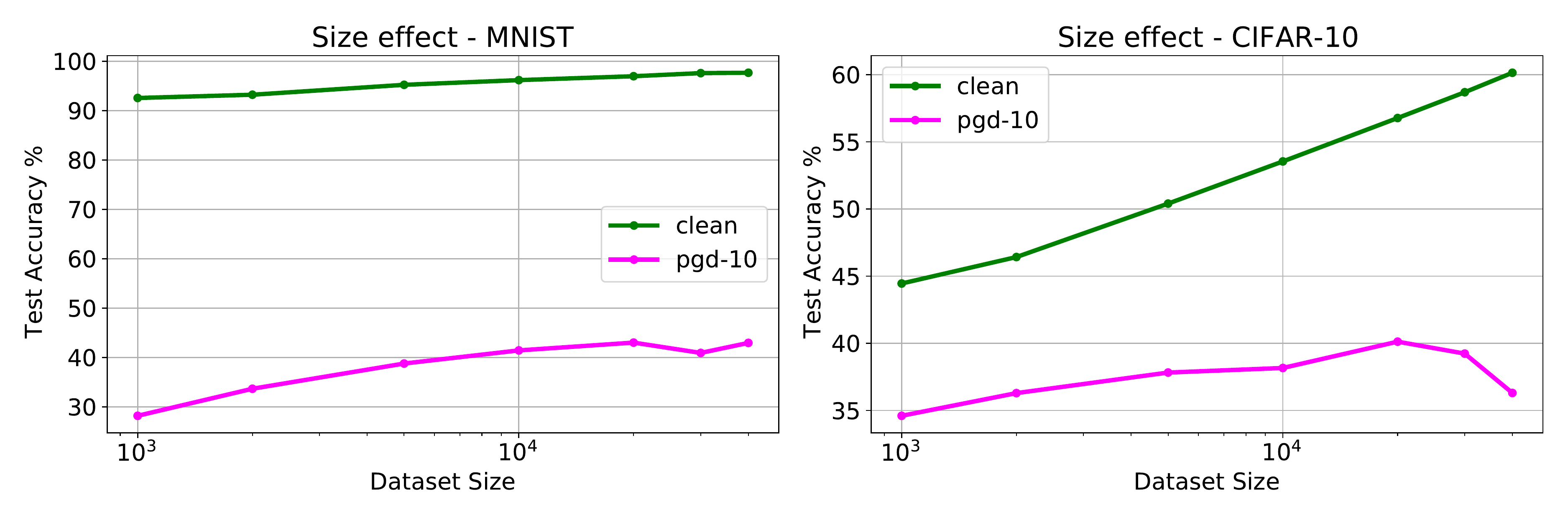}
    \caption{Performance of \texttt{adv-KIP} outputs as a function of dataset size. Setting: \texttt{FC3} kernel, $|\mathcal{X}_T| = 10k$, $\ell_\infty$ adversary, $\epsilon=0.3$ ($\epsilon=8/255$), $\alpha=0.1$ ($\alpha=2/255$).}
    \label{fig:scale}
\end{figure}

We also examine the importance of scale on the final robustness of the dataset. We run \texttt{adv-KIP} on MNIST and CIFAR-10, using an \texttt{FC3} kernel for a wide variety of different dataset sizes $|\mathcal{X}_S|$, while keeping the rest of the hyperparameters the same. We optimize for 500 meta epochs and use early stopping if robust validation accuracy ceases to improve. For all the runs, we observe that we stop early, so we conclude that all the datasets were given a fair chance. On the test set we then evaluate the same kernel classifier fit with the optimized data, recording clean and robust accuracy. The scaling laws are shown in Fig \ref{fig:scale}. We observe diminishing returns on the performance of the models, after we exceed some required sample size. For the more challenging setting (CIFAR-10) it is the clean accuracy that sees the greater increase. The fact that the robust curve ``bends" in the end might mean that we could optimize more agressively by harming clean accuracy (without implementing early stopping).

\subsection{Neural Networks results}

In this section, we study whether datasets produced with \texttt{adv-KIP} can be used to train robust neural networks.

\textbf{Wide fully connected neural networks.} As a first step, we perform experiments on the same architecture as in the NTK used in the meta-optimization of the dataset. We consider fully connected ReLU models with 2, 3, 5 and 7 layers of width $1024$ and train them using Adam. We refer to App. \ref{app:expdetails} and \ref{sec:wide} for full experimental details and results. We find that in all cases the networks learn successfully with clean and robust test performance that matches or exceeds those of the kernels (even when evaluating against stronger attacks with more PGD steps than used during the meta-optimization). We view this is an indication that features learned by kernels are surprisingly relevant for (robust) classification in neural networks.


\textbf{Robustness of Common Architectures.} We now turn our attention to commonly employed convolutional neural networks to study the relevance of our datasets for robust classification using modern architectures. In particular we report how datasets generated using \texttt{FC} kernels transfer to such convolutional architectures, since the small datasets generated from \texttt{Conv} kernels were challenging to fit with neural networks. We consider a simple CNN with 3 convolutional layers with max-pooling operations on MNIST, and AlexNet \cite{AlexNet}, VGG11 \citep{Simonyan15} and ResNet20 \citep{he2016deep} on CIFAR-10. We perform a sweep over different learning rates and training algorithms (gradient descent, gradient descent with weight decay and Adam). Experimental details can be found in App.~\ref{app:expdetails}.


Tables \ref{tab:nn-modern-mnist-new} and 
\ref{tab:adv_KIP_and_AT} summarize our findings. We note an astonishing ``boost" in robust test accuracy against gradient-based attacks of these convolutional networks (in all models but ResNet\footnote{We currently do not have an explanation on why ResNet performs so poorly in terms of robust accuracy. We hypothesize that the model performs feature learning that is entirely different from the other models and is not efficient on features that are relevant for an \texttt{FC} kernel.}) when compared to the previous results on kernels and fully connected networks. 
Very remarkably, it seems that datasets optimized for relatively simple kernels ``transfer" their PGD-performance to networks far removed from the idealistic infinite width regime, even to more expressive architectures. We also note that the results are robust to the initialization of the networks, as evidenced by the deviations reported in the tables.

\begin{table}[h]
    \caption{MNIST with simple-CNN: Test accuracies when trained on \texttt{adv-KIP} datasets optimized with \texttt{FC} kernels (first 3 rows). We also show test accuracies for the adversarially trained simple-CNN (without any data augmentation). AA refers to the {\em AutoAttack} test suite. Setting: $\ell_\infty$, $\epsilon=0.3, \alpha=0.1$.}
    \label{tab:nn-modern-mnist-new}
    \centering
        \resizebox{\columnwidth}{!}{
    \begin{tabular}{ccccc}
        \toprule
        & & \multicolumn{3}{c}{\textbf{Robust}} \\
        \textbf{Dataset} & Clean & FGSM & PGD40 & AA \\
        \midrule
        \texttt{FC3} & 98.15 $\pm$ 0.12 & 98.06 $\pm$ 0.18 & 97.17 $\pm$ 0.10  & \multicolumn{1}{c}{0.00 $\pm$ 0.00} \\
        \texttt{FC5} & \multicolumn{1}{c}{97.96 $\pm$ 0.55}                       & \multicolumn{1}{c}{97.87 $\pm$ 0.64} & \multicolumn{1}{c}{97.20 $\pm$ 0.74} & \multicolumn{1}{c}{0.00 $\pm$ 0.00} \\
        \texttt{FC7} & \multicolumn{1}{c}{98.03 $\pm$ 0.16} & \multicolumn{1}{c}{97.91 $\pm$ 0.22} & \multicolumn{1}{c}{97.14 $\pm$ 0.43} & \multicolumn{1}{c}{0.00 $\pm$ 0.00} \\
        \midrule
        \multicolumn{1}{c}{Adversarial Training}   & \multicolumn{1}{c}{99.11}      & \multicolumn{1}{c}{97.52} & \multicolumn{1}{c}{95.82} & \multicolumn{1}{c}{88.77} \\
        \bottomrule
    \end{tabular}}
\end{table}

\begin{table}[h]
    \caption{CIFAR-10: Test accuracies of several convolutional architectures trained on an \texttt{adv-KIP} dataset produced from an \texttt{FC3} kernel of $|\mathcal{X}_S|=40,000$ (upper), and Adversarial Training baselines (lower). Setting: $\ell_\infty$, $\epsilon=8/255, \alpha=2/255$. For further results on smaller attack radii, see Table \ref{tab:AA_smallR_AdvKIP} in App. \ref{app:small_eps}.}
    \label{tab:adv_KIP_and_AT}
    \centering
        \resizebox{\columnwidth}{!}{
    \begin{tabular}{ccccc}
        
        \toprule
        
        \multicolumn{5}{c}{ \texttt{adv-KIP}}\\ 
        \textbf{Neural Net} & Clean & FGSM & PGD20 & AA \\ 
        \midrule
        
        \multicolumn{1}{c}{Simple CNN}    & \multicolumn{1}{c}{72.10 $\pm$ 0.10}     & \multicolumn{1}{c}{67.45 $\pm$ 0.37}                       & \multicolumn{1}{c}{67.03 $\pm$ 0.24} & \multicolumn{1}{c}{0.00 $\pm$ 0.00} \\ 
\multicolumn{1}{c}{AlexNet}       & \multicolumn{1}{c}{68.87 $\pm$ 0.76}     & \multicolumn{1}{c}{49.30 $\pm$ 0.69}                       & \multicolumn{1}{c}{49.06 $\pm$ 0.63} & \multicolumn{1}{c}{0.89 $\pm$ 1.41} \\ 
\multicolumn{1}{c}{VGG11} & \multicolumn{1}{c}{74.88 $\pm$ 0.45}     & \multicolumn{1}{c}{53.98 $\pm$ 9.71}      & \multicolumn{1}{c}{53.18 $\pm$ 10.32} & \multicolumn{1}{c}{0.27 $\pm$ 0.18} \\ 
\multicolumn{1}{c}{ResNet20} & \multicolumn{1}{c}{81.53 $\pm$ 0.59}     & \multicolumn{1}{c}{4.82 $\pm$ 1.45}      & \multicolumn{1}{c}{0.20 $\pm$ 0.16} & \multicolumn{1}{c}{0.00 $\pm$ 0.00}  \\
\hline
\multicolumn{5}{c}{Adversarial Training Baseline}\\ 
\textbf{Neural Net} & Clean & FGSM & PGD20 & AA \\
\midrule
\multicolumn{1}{c}{Simple CNN} & \multicolumn{1}{c}{58.07} & 
\multicolumn{1}{c}{33.94} & 
\multicolumn{1}{c}{31.49} &
\multicolumn{1}{c}{26.18}
\\
\multicolumn{1}{c}{AlexNet} & \multicolumn{1}{c}{44.35} &                   \multicolumn{1}{c}{30.12} & 
\multicolumn{1}{c}{24.41}     & 
\multicolumn{1}{c}{18.95}  \\
\multicolumn{1}{c}{VGG11} & \multicolumn{1}{c}{69.69}      & \multicolumn{1}{c}{31.41}                 & \multicolumn{1}{c}{24.88}   &   \multicolumn{1}{c}{24.09} \\
\multicolumn{1}{c}{ResNet20} & \multicolumn{1}{c}{73.95}      & \multicolumn{1}{c}{46.37} & \multicolumn{1}{c}{39.17}   & 
\multicolumn{1}{c}{35.35} \\
\bottomrule
    \end{tabular}}
\end{table}

Fig.~\ref{fig:mega_cnn} (top left) shows accuracy curves for the CNN when trained on the optimized CIFAR-10 dataset. While accuracy on the train (optimized) data together with the clean test accuracy increase rapidly, it is only after 250 epochs that robust accuracy starts to increase (after we hit 0 training loss). We hypothesize that this might be due to the fact that \texttt{adv-KIP} optimizes using the expression that corresponds to the {\em end} of training in neural networks. 

It seems promising at first that modern networks trained with \texttt{adv-KIP} datasets without much tuning enjoy astonishing defense properties against PGD-attacks in various settings, similar, or in some cases even higher, than what truly robust models (i.e adversarially trained) obtain.
However, as can be seen in the last column of Tables \ref{tab:nn-modern-mnist-new} and \ref{tab:adv_KIP_and_AT}, when we evaluate their robustness with the adaptive methods of AutoAttack \citep{croce2020reliable}, we observe a sharp drop in the performance (close to 0\% in all cases). The purpose of the AutoAttack benchmark, which includes 4 different attacks, some of which do not use gradient information, is to provide a minimal adaptive attack suite to uncover shortcomings in a defense. See Table \ref{tab:AA-decomp} in App. \ref{app:aa_decomp} for a breakdown of robustness by AutoAttack components.

We find that datasets produced with \texttt{adv-KIP} suffer from what is commonly termed the \textit{obfuscated gradient} phenomenon \citep{Ath+18}, a situation where model gradients do not provide good directions for generating successful adversarial examples. However, in the past, this has only been observed with techniques that were either introducing non-differentiable parts in the inference pipeline or stochasticity to the model. Interestingly, we now observe this phenomenon from altering the {\em training}  data alone and, even more remarkably, from data optimized using kernels.

To check whether this is a shortcoming of our optimization method or a more general phenomenon related to "robustified" datasets, we turn our attention next to the only other method we are aware of that provides some notable robustness via standard training \citep{Ily+19}.

\subsection{Non-robustness of Robust Features dataset}\label{sec:rfd-experiments}

To illustrate the theory of the presence of robust and non-robust features in the data, \citet{Ily+19} have introduced a ``robustified" data set (termed \textit{Robust Features Dataset} or RFD from now on), which was sufficient to ensure robust predictions on the test set by standard training only. RFD is generated by traversing the representation layer of an adversarially trained neural network, and was thus believed to provide a general sense of robustness \citep{Ily+19}. More specifically, given a mapping $x \mapsto g(x)$ of an input $x$ to the penultimate (``representation") layer of an adversarially trained neural net, a ``robustified" input is obtained by optimizing $\min_{x_r} \|g(x)-g(x_r)\|_2$, starting from a random data point using gradient descent, thus enforcing that the robust representations of $x$ and $x_r$ are similar; while $x_r$ does not contain non-robust features given a starting point that is uncorrelated with the label of $x$.

We use an $\ell_\infty$-adversarially trained ResNet50 to generate an RFD version of CIFAR-10\footnote{Note that the {\em publicly} available dataset of \cite{Ily+19} is derived from an adversarially trained network trained against an $\ell_2$ adversary, so for completeness we include an $\ell_2$ evaluation of that dataset in App. \ref{app:publicMadry}, where the findings are similar.}. To ensure a fair comparison with our methods, we train the same set of models as in the previous subsection. Table \ref{tab:madry_infty} contains the results.

\begin{table}[]
\caption{Test accuracies for various models trained on a 50K $\ell_\infty$ Robust Features dataset \citep{Ily+19}) for CIFAR-10. Setting: $\ell_\infty$, $\epsilon=8/255, \alpha=2/255$. For further results on smaller attack radius, see Table \ref{tab:AA_smallR_RFDinf} in App. \ref{app:small_eps}.}
\label{tab:madry_infty}
\centering
\resizebox{\columnwidth}{!}{
\begin{tabular}{|cccc|}
\toprule
\multicolumn{4}{c}{\begin{tabular}[c]{@{}c@{}} Robust Features dataset \citep{Ily+19}\end{tabular}}                                     \\ \midrule
\multicolumn{1}{c}{\textbf{Neural Net}} & \multicolumn{1}{c}{Clean} & \multicolumn{1}{c}{PGD20} &  \multicolumn{1}{c}{AA} \\ \midrule
\multicolumn{1}{c}{Simple CNN} & \multicolumn{1}{c}{59.15 $\pm$ 0.37} & \multicolumn{1}{c}{52.91 $\pm$ 0.66} &  \multicolumn{1}{c}{0.00 $\pm$ 0.00} \\
\multicolumn{1}{c}{AlexNet}       & \multicolumn{1}{c}{51.62 $\pm$ 1.14} & \multicolumn{1}{c}{25.64 $\pm$ 4.32} &   \multicolumn{1}{c}{0.02 $\pm$ 0.03} \\
\multicolumn{1}{c}{VGG11} & \multicolumn{1}{c}{61.59 $\pm$ 0.80} & \multicolumn{1}{c}{34.64 $\pm$ 8.47} &  \multicolumn{1}{c}{0.40 $\pm$ 0.42} \\ 
\multicolumn{1}{c}{ResNet20}       & \multicolumn{1}{c}{66.29 $\pm$ 0.70} & \multicolumn{1}{c}{7.35 $\pm$ 3.10} & \multicolumn{1}{c}{0.00 $\pm$ 0.00} \\ \bottomrule
\end{tabular}}
\end{table}

Confirming the findings of \citet{Ily+19}, we see that the trained models record high robustness against PGD attacks. However, all of the models \textit{fail} to defend against AutoAttack. This is a surprising finding, since the dataset was generated using adversarially trained networks that guarantee a wide sense of robustness \citep{Cro+20}. In Fig. \ref{fig:mega_cnn} (middle), we can see that the simple CNN when trained on the RFD exhibits remarkably similar accuracy curves as when being trained on an \texttt{adv-KIP} dataset. In the next subsection, we dive deeper into the similarities between the two classes of datasets.

\subsection{Overconfidence gives a false sense of security}\label{sec:shortcomings}

\begin{figure*}[h]
    \centering
    \includegraphics[scale=0.24]{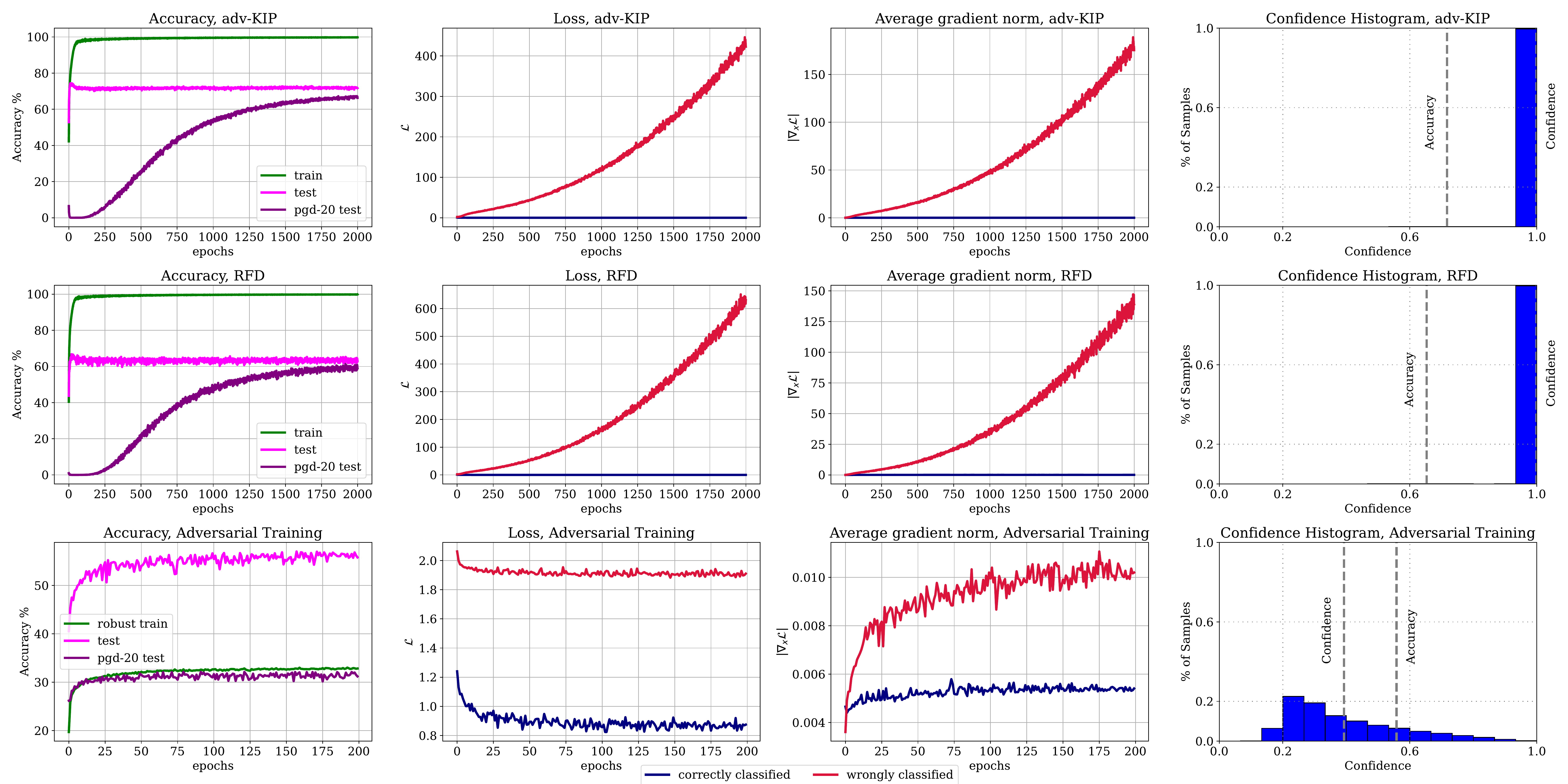}
    \caption{Several statistics of CNN models trained on an \texttt{adv-KIP} dataset (\textbf{first row}), an RFD (\textbf{second row}), and trained with adversarial training (\textbf{third row}). From Left to Right: Accuracy, Test Loss (decomposed in correctly and wrongly classified examples); Average Gradient Norm of Test data; Confidence Histogram. Base dataset: CIFAR-10.}
    \label{fig:mega_cnn}
\end{figure*}

We analyze properties of models that were trained with either an \texttt{adv-KIP} or an RF dataset and yield superficial robustness. We discover many common properties and signatures of failure, and contrast them to adversarially trained neural networks. These similarities might be germane to all attempts that explicitly optimize the dataset with gradient based approaches. 

Fig.~\ref{fig:mega_cnn} (second and third column) shows the loss value and the gradient magnitude of the models throughout training, where we further decomposed the metrics into 2 parts; one coming from correctly classified images and one from misclassified ones. We observe that for both datasets the loss increases on the misclassified examples, concurrently with an increase of the average norm of the gradient. In contrast, for correctly classified examples we see both quantities progressively vanish. This behavior, together with the false sense of robustness that AutoAttack evaluation reveals, suggests that the model learns to shatter the gradients locally in the neighborhood of correctly classified examples, causing simple gradient-based attacks to fail. We find that during our distillation procedure this is indeed the case (Fig.~\ref{fig:avg_grad_advkip} in the Appendix), and the data optimization effectively shrinks the gradients of the model.

In other words, the model essentially ``sacrifices" performance on the subset of data that is misclassified, in order to fit the rest in a pseudo robust way. This is in sharp contrast with the situation during adversarial training, where the loss decreases on all data simultaneously (Fig. \ref{fig:mega_cnn}, third row). This is already an indication of failure of learning. Interestingly, Fig. \ref{fig:mega_cnn} (third column) shows that this failure of learning is also evident in the gradient norms of the loss with respect to the \textit{input}. The average gradient norm on the wrongly classified points explodes with the number of epochs for both the \texttt{adv-KIP} dataset and the RFD. 

Finally, we study the calibration of the models. In particular, we compute confidence histograms and reliability diagrams \citep{guo2017calibration} for the 3 models.  As Fig. \ref{fig:mega_cnn} (last column) shows, the models that were trained with the \texttt{adv-KIP} dataset or the RFD are extremely confident on nearly all test examples, being poorly calibrated, while the adversarially trained model has a completely different, more balanced, profile (it is less confident than accurate). We refer to Fig. \ref{fig:reliability_cnn} in App. \ref{app:confidence} for the reliability diagram of the 3 models that provides further evidence for the above. In a sense, this overconfidence on the samples is what makes the gradients non-informative. 

Understanding why the solution of Eq.~(\ref{eq:data_framework}) (data level) with gradient based methods gives a false sense of robustness, while the solution of Eq.~(\ref{eq:rob_opt}) (parameters level) also with gradient based methods induces more widespread robustness seems a very interesting and challenging open question for future work to address.


\section{Conclusion}

In this work, we approached the problem of robust classification in machine learning from a data-based point of view. We introduced a meta-learning algorithm, \texttt{adv-KIP}, which produces datasets that furnish a great range of classifiers with notable robustness against gradient-based attacks. However, our analysis demonstrates that this robustness is rather fallacious and the models essentially learn representations that provide no meaningful gradients. Quite remarkably, when we revisited seminal prior work \citep{Ily+19} that claimed standard training on a robust dataset can be an actual defense versus adversarial attacks, we found the same pathological situation. We believe our findings provide many new insights on the role of data in adversarially robust classification and should be helpful for potential new data-based approaches in the future.

\bibliography{icml2023}
\bibliographystyle{icml2023}


\newpage
\appendix
\onecolumn
\section{Appendix}




\subsection{Related Work}\label{app:previouswork}

\textbf{Distributionally Robust Optimization and Adversarial Augmentation.} Related to our work are also works on distributionally robust optimization \citep{Sin+18} and adversarial data augmentation for out-of-distribution generalization \citep{Vol+18}. The latter proposes an algorithm that augments the training dataset \textit{on-the-fly} (i.e. during training of a neural net) with worst-case samples from a target distribution. In contrast, our method optimizes the original dataset against worst-case samples/adversarial examples from the original distribution, which correspond to a final predictor (kernel machine). The only prior work  that gives an algorithm based on  NTK theory to derive dataset perturbations in some adversarial setting is due to \citet{YuWu21}, yet with entirely different focus. It deals with what is coined {\em generalization attacks}: the process of altering the training data distribution to prevent models to generalise on clean data. To our knowledge, KIP \citep{Ngu+21a,Ngu+21b} and this NTGA algorithm are the only examples of leveraging NTKs for dataset optimization.

\subsection{KIP baseline}\label{app:KIP}

The original KIP algorithm \citep{Ngu+21a,Ngu+21b} is designed to reduce the size of the training set, while keeping the induced accuracy close to the original one. It could be reasonable to hypothesize that such  information compression might possibly lead to an increase of robustness as well. As a sanity check we evaluate the robust accuracy of these datasets. We also produce larger dataset using the KIP algorithm (of the same size as used in our \texttt{adv-KIP} experiments). Table \ref{tab:kip_repro} shows that effectively the robustness of the datasets remains close to 0, as is the case for the original datasets.
This indicates the clear need to adjust the optimization objective to robust performance, as is done in the \texttt{adv-KIP} algorithm.

\begin{table}[h]
    \caption{KIP baseline datasets (reproduced). Setting: No preprocessing/data augmentation, $|\mathcal{X}_T|=$1,000 images, learned labels, mse loss, lr=$1$e-3, datasets were optimized for 1000 epochs, with potential early stopping if validation accuracy did not increase across 200 epochs. Random seed denotes different draws of the initial support images. Standard setting for robust evaluation: $\epsilon=0.3$ and $\epsilon=8/255$, respectively.}
    \label{tab:kip_repro}
    \centering
    \begin{tabular}{ccccc}
        \toprule
        & \textbf{Kernel}, $|\mathcal{X}_s|$ & Clean & \textbf{FGSM} \\
        \midrule
        \multirow{2}{*}{MNIST} & FC3, 5k & 97.51 $\pm$ 0.03 & 0.00 $\pm$ 0.00 \\
        & FC7, 30k & 98.23 $\pm$ 0.06 & 0.00 $\pm$ 0.00 \\
        \midrule
        \multirow{3}{*}{CIFAR-10} & FC3, 1k & 48.45 $\pm$ 0.34 & 2.50 $\pm$ 0.21 \\
        & FC3, 5k & 52.48 $\pm$ 0.23 & 0.22 $\pm$ 0.05 \\
        & FC3, 10k & 54.04 $\pm$ 0.41 & 0.10 $\pm$ 0.04 \\
        \bottomrule
    \end{tabular}
\end{table}

We also evaluated \texttt{FC$\{3, 5, 7\}$} and \texttt{Conv$\{3, 5, 7\}$} kernels (together with a Convolutional Kernel with 1 hidden layer followed by global average-pooling) on  datasets (with 50 images per class) released by \citep{Ngu+21b} and we found their FGSM robustness to be 0$\%$ in all cases. URLs for the  datasets we considered: \href{gs://kip-datasets/kip/mnist/ConvNet_ssize500_nozca_l_noaug_ckpt1000.npz}{1st} and \href{gs://kip-datasets/kip/mnist/ConvNet_ssize500_nozca_l_noaug_ckpt50000.npz}{2nd}.


\subsection{Experimental Details for Neural Network experiments}\label{app:expdetails}

For all models trained on an \texttt{adv-KIP} dataset or an RFD, we use the Adam optimizer and perform a small grid search for the learning rate, picking the best model with respect to robust accuracy.

On MNIST, we train fully connected networks of width 1024 in Sec.~\ref{sec:wide}, and the simple-CNN network in Sec.~\ref{sec:experiments}. FC networks are trained for 2,000 epochs and the simple-CNN network for 800 epochs.

On CIFAR-10, we again train fully connected networks of width 1024 in Sec.~\ref{sec:wide}, and the simple-CNN, AlexNet and VGG11 networks in Sec.~\ref{sec:experiments}. We train all these networks for 2,000 epochs.

For the Adversarial Training baseline, on MNIST, we adopt the setting of \cite{Mad+18}, that is we train the simple-CNN network with the Adam optimizer towards convergence, and set the initial learning rate to 1e-4. In \cite{Mad+18} the number of epochs was set to 100, while we use 200. On CIFAR-10, since we do not use data augmentation, we train with both SGD and Adam for 200 epochs for each model, and pick the better one in terms of robustness. For the simple-CNN and AlexNet, the Adam optimizer is better. For VGG11, we use the SGD optimizer, with initial learning rate 1e-1, decay rate of 10 at the 100-th and the 150-th epoch, and with weight decay 5e-4.


\textbf{Simple-CNN architecture:} 
We use a simple convolutional architecture with three convolutional layers and a linear layer. Each convolutional layer computes a convolution with a 3$\times$3 kernel, followed by a ReLU and a max-pooling layer (of kernel size 2$\times$2 and stride 2). The linear layer is fully-connected  with ten outputs. All convolutional layers have a fixed width of 64.

\textbf{Training of Convolutional Nets:}
  We use the Adam optimizer \citep{KiBa15} and perform a small grid search over the fixed learning rate. We stop training when robust validation accuracy ceases to decrease, where we measure against PGD40 attacks for MNIST and PGD20 attacks for CIFAR-10, as is often standard. We report the best results across the sweep for FGSM and PGD test accuracies. After picking the best learning rate, for each experiment in this paper, we report the mean and standard deviation of three experiments with different seeds.

\textbf{Description of Evaluation Metrics:}
For all the adversarial attack related measurements including FGSM, $\ell_\infty$ PGD and $\ell_2$ PGD, we adopt the CleverHans code implementation \citep{Pap+18}. For $\ell_\infty$ PGD, on MNIST we use step size 0.1 and radius 0.3, while on CIFAR-10 we use step size 2/255 and radius 8/255. For $\ell_2$ PGD on CIFAR-10, we use step size 15/255 and radius 128/255.

For AutoAttack, we adopt the open-source original implementation \citep{croce2020reliable,croce2021mind}. As for the other attacks, we set $\epsilon=0.3$  for MNIST and 8/255 for CIFAR-10 for $\ell_\infty$ attacks and adopt the 128/255 radius for CIFAR-10 for the $\ell_2$ adversary.

\subsection{Transfer Results to Wide FC Networks} \label{sec:wide}

Here, we evaluate how well datasets produced with kernel methods in Algorithm \ref{alg:adv-kip} transfer to relatively wide neural nets of the same architecture and depth as the NTKs used in the \texttt{adv-KIP} optimization.  We implement multilayer fully connected neural nets of width 1024 and perform a hyperparameter search for the (constant) learning rate. We use the Adam optimizer \citep{KiBa15} and test for both FGSM and PGD accuracy, where we apply the most common PGD attacks (PGD40 for MNIST and PGD20 for CIFAR-10). Table \ref{tab:kernel-to-nn:merge} summarize our results.

\begin{table}[h]
    \caption{Transferability : Kernel to Neural Network of same architecture, test accuracy in $\%$. For MNIST, we test with PGD-40, and for CIFAR, we test with PGD-20.}
    \label{tab:kernel-to-nn:merge}
    \centering
    \begin{tabular}{ccccc}
        \toprule
        & & & \multicolumn{2}{c}{\textbf{Robust}} \\
        \textbf{Dataset} & \textbf{Kernel}, $|\mathcal{X}_s|$ & Clean & FGSM & PGD \\
        \midrule
        \multirow{3}{*}{MNIST} & FC3, 30k & 80.08 & 77.67 & 53.85  \\
        & FC5, 30k & 97.75 & 64.83 & 35.14 \\
        & FC7, 30k & 97.45 & 70.58 & 40.70 \\
        \midrule
        \multirow{2}{*}{CIFAR-10} & FC2, 40k & 46.29 & 20.98 & 16.89  \\
        & FC3, 40k & 46.33 & 40.07 & 39.15 \\
        \bottomrule
    \end{tabular}
\end{table}



We find that robustness properties transfer well from kernels to their corresponding neural networks. Our sweeps also show that this holds for a rather wide range of learning rates, evidencing a certain insensitivity to exact parameter choices.

\subsection{Smaller Radius AutoAttack Results}\label{app:small_eps}
To further investigate, we deploy a fine-grained series of smaller radius AutoAttack to the models trained with our \texttt{adv-KIP} dataset and the $\ell_\infty$ RF Dataset. The test results are shown in Table \ref{tab:AA_smallR_RFDinf} and \ref{tab:AA_smallR_AdvKIP}. 
Our \texttt{adv-KIP} dataset shows consistently better PGD and AA robustness across multiple models.

\begin{table}[]
\caption{AutoAttack Test accuracies for various models trained on the $\ell_\infty$ RFD.}
\label{tab:AA_smallR_RFDinf}
\centering
\begin{tabular}{|ccccc|}
\toprule
\multicolumn{5}{c}{\begin{tabular}[c]{@{}c@{}}\textbf{CIFAR-10 AutoAttack Accuracy, Smaller Radius} \end{tabular}}                                     \\ \midrule
\multicolumn{1}{c}{\textbf{Neural Net}} 
& \multicolumn{1}{c}{AA $\ell_\infty$ 4/255} & \multicolumn{1}{c}{AA $\ell_2$ 64/255}                 &  \multicolumn{1}{c}{AA $\ell_\infty$} & \multicolumn{1}{c}{AA $\ell_2$}\\ \midrule
\multicolumn{1}{c}{Simple CNN}    
& \multicolumn{1}{c}{0.02 $\pm$ 0.01} & \multicolumn{1}{c}{4.91 $\pm$ 0.29} &  \multicolumn{1}{c}{0.00 $\pm$ 0.00} & \multicolumn{1}{c}{0.06 $\pm$ 0.03}   \\
\multicolumn{1}{c}{AlexNet}       
& \multicolumn{1}{c}{0.98 $\pm$ 0.15} & \multicolumn{1}{c}{5.98 $\pm$ 0.78} &   \multicolumn{1}{c}{0.02 $\pm$ 0.03} & \multicolumn{1}{c}{0.37 $\pm$ 0.13}   \\
\multicolumn{1}{c}{VGG11} 
& \multicolumn{1}{c}{3.92 $\pm$ 1.76} & \multicolumn{1}{c}{19.85 $\pm$ 3.31} &  \multicolumn{1}{c}{0.40 $\pm$ 0.42} & \multicolumn{1}{c}{4.71 $\pm$ 2.05}   \\ 
\multicolumn{1}{c}{ResNet20}       
& \multicolumn{1}{c}{0.02 $\pm$ 0.02} & \multicolumn{1}{c}{4.12 $\pm$ 0.87} &   \multicolumn{1}{c}{0.00 $\pm$ 0.00} &\multicolumn{1}{c}{0.05 $\pm$ 0.04}      \\ \bottomrule
\end{tabular}
\end{table}

\begin{table}[h]
\caption{AutoAttack Test accuracies for various models trained on our \texttt{adv-KIP} Dataset.}
\label{tab:AA_smallR_AdvKIP}
    \centering
    \begin{tabular}{ccccc}
        
        \hline
        
        \multicolumn{5}{c}{\textbf{CIFAR-10 AutoAttack Accuracy, Smaller Radius}}\\ 
        \textbf{Neural Net} 
        & AA $\ell_\infty$ 4/255 & AA $\ell_2$ 64/255 & AA $\ell_\infty$ & AA $\ell_2$\\ 
        \hline
        
        \multicolumn{1}{c}{Simple CNN}    
        & \multicolumn{1}{c}{0.02 $\pm$ 0.00} & \multicolumn{1}{c}{7.48 $\pm$ 0.20} & \multicolumn{1}{c}{0.00 $\pm$ 0.00} & \multicolumn{1}{c}{0.14 $\pm$ 0.01} \\ 
\multicolumn{1}{c}{AlexNet}       
& \multicolumn{1}{c}{4.98 $\pm$ 2.32} & \multicolumn{1}{c}{19.27 $\pm$ 1.81} & \multicolumn{1}{c}{0.89 $\pm$ 1.41} & \multicolumn{1}{c}{3.94 $\pm$ 2.65} \\ 
\multicolumn{1}{c}{VGG11} 
& \multicolumn{1}{c}{6.20 $\pm$ 0.76}& \multicolumn{1}{c}{31.41 $\pm$ 1.85} & \multicolumn{1}{c}{0.27 $\pm$ 0.18} & \multicolumn{1}{c}{8.56 $\pm$ 1.00} \\ 
\multicolumn{1}{c}{ResNet20} 
& \multicolumn{1}{c}{0.00 $\pm$ 0.00}& \multicolumn{1}{c}{0.69 $\pm$ 0.04} & \multicolumn{1}{c}{0.00 $\pm$ 0.00} & \multicolumn{1}{c}{0.00 $\pm$ 0.00}  \\
\bottomrule
    \end{tabular}
\end{table}

\subsection{AutoAttack Suite Decomposition Analysis}\label{app:aa_decomp}
In Table \ref{tab:AA-decomp}, we decompose the AA test suite into its individual components and evaluate them independently on models trained on our Adv-KIP dataset, the $\ell_2$ RFD dataset, and the $\ell_\infty$ RFD dataset. A clear trend of high APGD-CE accuracy and low other accuracies emerges. This corroborates our discussion that the fake robustness arises from the dynamics of overconfidence that operates on the logits to make the gradient vanish.

\begin{table}[h]
\caption{Test Accuracy on CIFAR-10 with the $\ell_\infty$ AA suite decomposition. $\epsilon=8/255.$}
\label{tab:AA-decomp}
\centering
\begin{tabular}{ccccccc}
\toprule
\multicolumn{7}{c}{\begin{tabular}[c]{@{}c@{}}\textbf{CIFAR-10 Accuracy with individual $\ell_\infty$ AutoAttack Components}\end{tabular}}                                     \\ \midrule
\multicolumn{1}{c}{\textbf{Dataset}} & \multicolumn{1}{c}{\textbf{Neural Net}} & \multicolumn{1}{c}{Clean} & \multicolumn{1}{c}{APGD-CE}                 & \multicolumn{1}{c}{APGD-T} & \multicolumn{1}{c}{FAB-T} & \multicolumn{1}{c}{SQUARE} \\ \midrule
\multirow{4}{*}{\texttt{adv-KIP}} & \multicolumn{1}{c}{Simple CNN}    & \multicolumn{1}{c}{72.10 $\pm$ 0.10} & \multicolumn{1}{c}{66.33 $\pm$ 0.21}                     & \multicolumn{1}{c}{0.00 $\pm$ 0.00} &   \multicolumn{1}{c}{0.00 $\pm$ 0.01}     &   \multicolumn{1}{c}{0.03 $\pm$ 0.01}   \\
& \multicolumn{1}{c}{AlexNet}       & \multicolumn{1}{c}{68.87 $\pm$ 0.76} & \multicolumn{1}{c}{44.55 $\pm$ 0.78} & \multicolumn{1}{c}{1.44 $\pm$ 2.03} &  \multicolumn{1}{c}{7.42 $\pm$ 1.14}    &   \multicolumn{1}{c}{5.41 $\pm$ 2.24}   \\
& \multicolumn{1}{c}{VGG11} & \multicolumn{1}{c}{74.88 $\pm$ 0.45} & \multicolumn{1}{c}{48.81 $\pm$ 9.90}                 & \multicolumn{1}{c}{0.63 $\pm$ 0.15} & \multicolumn{1}{c}{6.33 $\pm$ 1.09}     &   \multicolumn{1}{c}{9.06 $\pm$ 1.67}   \\
& \multicolumn{1}{c}{ResNet20} & \multicolumn{1}{c}{81.53 $\pm$ 0.59} & \multicolumn{1}{c}{0.36 $\pm$ 0.05}                 & \multicolumn{1}{c}{0.00 $\pm$ 0.00} & \multicolumn{1}{c}{0.00 $\pm$ 0.00}      &   \multicolumn{1}{c}{0.01 $\pm$ 0.01}  \\ \midrule

\multirow{4}{*}{\textbf{$\ell_\infty$ RFD}} & \multicolumn{1}{c}{Simple CNN}    & \multicolumn{1}{c}{59.15 $\pm$ 0.37} & \multicolumn{1}{c}{52.18 $\pm$ 0.47}                     & \multicolumn{1}{c}{0.00 $\pm$ 0.00} &   \multicolumn{1}{c}{0.00 $\pm$ 0.00}     &   \multicolumn{1}{c}{0.03 $\pm$ 0.01}   \\
& \multicolumn{1}{c}{AlexNet}       & \multicolumn{1}{c}{51.62 $\pm$ 1.14} & \multicolumn{1}{c}{21.36 $\pm$ 4.10} & \multicolumn{1}{c}{0.10 $\pm$ 0.17} &  \multicolumn{1}{c}{1.56 $\pm$ 1.87}    &   \multicolumn{1}{c}{0.93 $\pm$ 0.18}   \\
& \multicolumn{1}{c}{VGG11} & \multicolumn{1}{c}{61.59 $\pm$ 0.80} & \multicolumn{1}{c}{29.72 $\pm$ 8.37}                 & \multicolumn{1}{c}{1.09 $\pm$ 1.15} & \multicolumn{1}{c}{4.39 $\pm$ 3.69}     &   \multicolumn{1}{c}{4.62 $\pm$ 2.32}   \\
& \multicolumn{1}{c}{ResNet20} & \multicolumn{1}{c}{66.29 $\pm$ 0.70} & \multicolumn{1}{c}{9.37 $\pm$ 1.67}                 & \multicolumn{1}{c}{0.00 $\pm$ 0.00} & \multicolumn{1}{c}{0.00 $\pm$ 0.00}      &   \multicolumn{1}{c}{0.10 $\pm$ 0.01} \\ \midrule

\multirow{4}{*}{\textbf{$\ell_2$ RFD}} & \multicolumn{1}{c}{Simple CNN}    & \multicolumn{1}{c}{65.25 $\pm$ 0.44} & \multicolumn{1}{c}{60.02 $\pm$ 0.39}                     & \multicolumn{1}{c}{0.00 $\pm$ 0.00} &   \multicolumn{1}{c}{0.00 $\pm$ 0.00}     &   \multicolumn{1}{c}{0.07 $\pm$ 0.05}   \\
& \multicolumn{1}{c}{AlexNet}       & \multicolumn{1}{c}{57.07 $\pm$ 1.25} & \multicolumn{1}{c}{21.42 $\pm$ 5.12} & \multicolumn{1}{c}{0.15 $\pm$ 0.23} &  \multicolumn{1}{c}{0.18 $\pm$ 0.09}    &   \multicolumn{1}{c}{0.93 $\pm$ 0.80}   \\
& \multicolumn{1}{c}{VGG11} & \multicolumn{1}{c}{68.41 $\pm$ 1.95} & \multicolumn{1}{c}{38.46 $\pm$ 10.44}                 & \multicolumn{1}{c}{0.95 $\pm$ 1.39} & \multicolumn{1}{c}{4.42 $\pm$ 3.08}     &   \multicolumn{1}{c}{6.31 $\pm$ 5.46}   \\
& \multicolumn{1}{c}{ResNet20} & \multicolumn{1}{c}{72.36 $\pm$ 0.15} & \multicolumn{1}{c}{0.19 $\pm$ 0.04}                 & \multicolumn{1}{c}{0.00 $\pm$ 0.00} & \multicolumn{1}{c}{0.00 $\pm$ 0.01}      &   \multicolumn{1}{c}{0.21 $\pm$ 0.07} \\ \bottomrule
\end{tabular}
\end{table}


\subsection{Robustness of the publicly available RFD}\label{app:publicMadry}

Here, we replicate the results of Sec. \ref{sec:rfd-experiments} for the publicly available RFD dataset of \cite{Ily+19}. Since it stems from a network trained against an $\ell_2$ adversary, we have also included evaluation against $\ell_2$-attacks. We refer to Table \ref{tab:nn-Madry} for the results. For all models, we observe a dramatic drop in their robustness once they are evaluated against AutoAttack.

\begin{table}[h]
\caption{Test accuracies for various models trained on the publicly-available 50K $\ell_2$ Robust Features dataset (RFD) for CIFAR-10. }
\label{tab:nn-Madry}
\centering
\resizebox{\textwidth}{!}{
\begin{tabular}{|cccccccc|}
\toprule
\multicolumn{8}{c}{\begin{tabular}[c]{@{}c@{}}\textbf{CIFAR-10 Accuracy with $\ell_2$ Robust Features dataset} \citep{Ily+19}\end{tabular}}                                     \\ \midrule
\multicolumn{1}{c}{\textbf{Neural Net}} & \multicolumn{1}{c}{Clean} & \multicolumn{1}{c}{PGD $\ell_\infty$ 20}                 & \multicolumn{1}{c}{PGD $\ell_2$ 20} & \multicolumn{1}{c}{AA $\ell_\infty$} & \multicolumn{1}{c}{AA $\ell_2$} & \multicolumn{1}{c}{AA $\ell_\infty$ 4/255} & \multicolumn{1}{c}{AA $\ell_2$ 64/255} \\ \midrule
\multicolumn{1}{c}{Simple CNN}    & \multicolumn{1}{c}{65.25 $\pm$ 0.44} & \multicolumn{1}{c}{60.73 $\pm$ 0.24}                     & \multicolumn{1}{c}{63.73 $\pm$ 0.40} &   \multicolumn{1}{c}{0.00 $\pm$ 0.00}     &   \multicolumn{1}{c}{0.47 $\pm$ 0.11}   & \multicolumn{1}{c}{0.09 $\pm$ 0.05} & \multicolumn{1}{c}{10.42 $\pm$ 0.60} \\
\multicolumn{1}{c}{AlexNet}       & \multicolumn{1}{c}{57.07 $\pm$ 1.25} & \multicolumn{1}{c}{25.12 $\pm$ 5.46} & \multicolumn{1}{c}{26.58 $\pm$ 4.80} &  \multicolumn{1}{c}{0.01 $\pm$ 0.02}    &   \multicolumn{1}{c}{0.62 $\pm$ 0.25} & \multicolumn{1}{c}{1.21 $\pm$ 0.52} & \multicolumn{1}{c}{8.11 $\pm$ 1.80}   \\
\multicolumn{1}{c}{VGG11} & \multicolumn{1}{c}{68.41 $\pm$ 1.95} & \multicolumn{1}{c}{42.92 $\pm$ 11.23}                 & \multicolumn{1}{c}{47.49 $\pm$ 6.12} & \multicolumn{1}{c}{1.19 $\pm$ 0.77}     &   \multicolumn{1}{c}{6.94 $\pm$ 2.47} & \multicolumn{1}{c}{4.65 $\pm$ 2.21} & \multicolumn{1}{c}{26.33 $\pm$ 3.00}   \\
\multicolumn{1}{c}{ResNet20} & \multicolumn{1}{c}{72.36 $\pm$ 0.15} & \multicolumn{1}{c}{0.08 $\pm$ 0.04}                 & \multicolumn{1}{c}{35.71 $\pm$ 3.04} & \multicolumn{1}{c}{0.00 $\pm$ 0.00}      &   \multicolumn{1}{c}{0.19 $\pm$ 0.06} & \multicolumn{1}{c}{0.05 $\pm$ 0.02} & \multicolumn{1}{c}{7.21 $\pm$ 1.44}   \\ \bottomrule
\end{tabular}
}
\end{table}

\subsection{Details on the Confidence and Reliability Visualization}\label{app:confidence}

\begin{figure}
    \centering
    \includegraphics[scale=0.17]{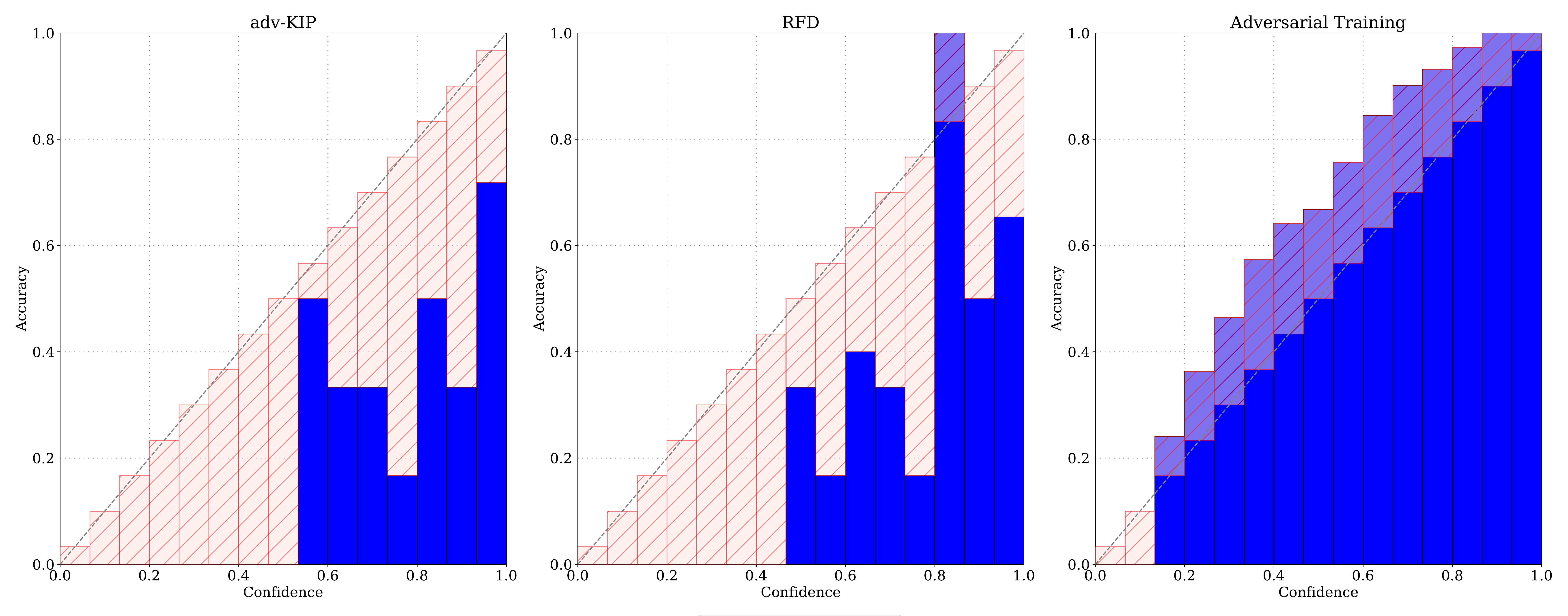}
    \caption{Relability Diagram for a Simple CNN trained on an \texttt{adv-KIP} dataset (left), an RFD (right), and optimized with adversarial training on CIFAR-10.}
    \label{fig:reliability_cnn}
\end{figure}

It has been shown that calibration of modern neural networks can be poor, despite advances in accuracy \citep{guo2017calibration, lakshminarayanan2017simple, wenzel2020hyperparameter, havasi2020training}. \cite{guo2017calibration} point out that more accurate and larger models tend to have worse calibration. A common measurement of miscalibration is the Expected Calibration Error (ECE) \citep{naeini2015obtaining}, which quantifies the difference in expectation between confidence and accuracy using binning. Since obtaining accurate estimation of ECE is difficult, due to the dependency of the estimator on the binning scheme, we adopt the reliability diagram \citep{degroot1983comparison, niculescu2005predicting} and the confidence histogram \citep{guo2017calibration}, both tools with nice visualization. In the confidence histogram, we display the distribution of the predicted confidence, \textit{i.e.,} the output probability of the predicted label, as a histogram. In the reliability diagram, we calculate the expected sample accuracy as a function of the confidence level by grouping all samples by their confidence. For a well-calibrated model, the reliability diagram should output the identity function, so we also plot the gap between the well-calibrated accuracy v.s. real accuracy. Fig. \ref{fig:reliability_cnn} shows the reliability diagram of Simple CNNs trained on an \texttt{adv-KIP} dataset (left), RFD (middle) and adversarially trained on CIFAR-10 (right). We notice how poorly calibrated are the models that were trained with the synthetic datasets, which is consistent with the findings of Fig. \ref{fig:mega_cnn} of the main text.

\subsection{Results using modifications of adv-KIP}\label{app:surrogate}

In this section, we present results where we modified the loss of either the inner or the outer loop of \texttt{adv-KIP}. 


In particular, for the inner loop, we replace the cross entropy with a rescaled version of the CW loss \citep{CaWa17}, called DLR loss \citep{croce2020reliable}. Let $z$ represent the pre-softmax logits. Recall that the cross entropy loss is defined as:
\begin{equation}
    \mathrm{CE}(x,y) = -z_y +\log(\sum_{i=1}^K e^{z_j})
\end{equation}

\citet{CaWa17} proposed to use the following (CW) loss for an attack:
\begin{equation}
    \mathrm{CW}(x,y) = -z_y + \max_{i\neq y} z_i
\end{equation}
We implement a variant of the above loss, namely the Difference of Logits Ratio (DLR) loss proposed in \citep{croce2020reliable}:
\begin{equation}
    \mathrm{DLR}(x,y) = -\frac{z_y - \max_{i\neq y} z_i}{z_{\pi_1} - z_{\pi_3}}
\end{equation}
where $\pi$ is the ordering of the components of $z$ in decreasing order (the untargeted version). This loss is invariant to scaling of the logits, and it has been used to detect cases where attacking the cross entropy loss fails due to overconfidence of the model.

For the outer loop,we incorporate the TRADES loss \citep{Zha+19} that aims to balance performance on clean and perturbed input points. Given a specific input $(x,y)$, TRADES optimizes over
\begin{equation}
    \mathcal L(f(x;\theta), y) + \lambda \max_{x'\in\mathcal B(x)} \mathcal L(f(x;\theta), f(x';\theta)),
\end{equation}
where $\lambda$ is a hyperparameter responsible for trading one accuracy for robustness (and vice versa).
In our case, the loss reads as
\begin{equation}
    \mathcal{L}(K_{\mathcal{X}_T \mathcal{X}_S} K_{\mathcal{X}_S \mathcal{X}_S}^{-1} \mathcal{Y}_S, \mathcal{Y}_T) + \lambda \max_{\mathcal X_{T'}\in\mathcal B(\mathcal X_T)} \mathcal L(K_{\mathcal{X}_T \mathcal{X}_S} K_{\mathcal{X}_S \mathcal{X}_S}^{-1} \mathcal{Y}_S, K_{\mathcal{X}_{T'} \mathcal{X}_S} K_{\mathcal{X}_S \mathcal{X}_S}^{-1} \mathcal{Y}_S).
\end{equation}




Tables \ref{tab:nn-modern-CW-40K} and \ref{tab:nn-modern-TRADES-40K} summarize our results on neural networks trained on the modified \texttt{adv-KIP} datasets with $\ell_\infty$ attacks. We do not see any notable deviation from the results that were presented in the main text.

\begin{table}[h]
    \centering
    \begin{tabular}{ccccc}
        \toprule
        & & \multicolumn{3}{c}{\textbf{Robust}} \\
        \textbf{Neural Net} & Clean & FGSM & PGD $\ell_\infty$ 20 & AA \\
        \midrule
        \multicolumn{1}{c}{Simple CNN}    & \multicolumn{1}{c}{70.87 $\pm$ 0.44}     & \multicolumn{1}{c}{65.06 $\pm$ 0.52}                       & \multicolumn{1}{c}{64.88 $\pm$ 0.51}& \multicolumn{1}{c}{0.00 $\pm$ 0.00}\\ 
\multicolumn{1}{c}{AlexNet}       & \multicolumn{1}{c}{63.58 $\pm$ 5.50}     & \multicolumn{1}{c}{47.62 $\pm$ 8.51}                       & \multicolumn{1}{c}{47.08 $\pm$ 8.65}& \multicolumn{1}{c}{0.11 $\pm$ 0.11}     \\
\multicolumn{1}{c}{VGG11} & \multicolumn{1}{c}{73.72 $\pm$ 1.90}     & \multicolumn{1}{c}{63.05 $\pm$ 4.14}      & \multicolumn{1}{c}{62.76 $\pm$ 4.40} & \multicolumn{1}{c}{2.11 $\pm$ 3.28}\\
        \bottomrule
    \end{tabular}
    \caption{Test accuracies of several convolutional architectures trained on an \texttt{adv-KIP} CIFAR-10 \textbf{DLR} dataset from the \texttt{FC3} kernel, with 40,000 samples. Setting: $\ell_\infty$, $\epsilon=8/255.$}
    \label{tab:nn-modern-CW-40K}
\end{table}

\begin{table}[h]
    \centering
    \begin{tabular}{ccccc}
        \toprule
        & & \multicolumn{3}{c}{\textbf{Robust}} \\
        \textbf{Neural Net} & Clean & FGSM & PGD $\ell_\infty$ 20 & AA \\
        \midrule
        \multicolumn{1}{c}{Simple CNN}    & \multicolumn{1}{c}{67.39 $\pm$ 0.18}     & \multicolumn{1}{c}{58.21 $\pm$ 0.33}                       & \multicolumn{1}{c}{58.03 $\pm$ 0.34} & \multicolumn{1}{c}{0.00 $\pm$ 0.00} \\ 
\multicolumn{1}{c}{AlexNet}       & \multicolumn{1}{c}{59.69 $\pm$ 1.13}     & \multicolumn{1}{c}{47.44 $\pm$ 8.00}                       & \multicolumn{1}{c}{47.08 $\pm$ 8.47}& \multicolumn{1}{c}{0.29 $\pm$ 0.17}     \\
\multicolumn{1}{c}{VGG11} & \multicolumn{1}{c}{68.31 $\pm$ 1.17}     & \multicolumn{1}{c}{60.97 $\pm$ 3.37}      & \multicolumn{1}{c}{60.61 $\pm$ 3.46}& \multicolumn{1}{c}{3.58 $\pm$ 2.01}\\
        \bottomrule
    \end{tabular}
    \caption{Test accuracies of several convolutional architectures trained on an \texttt{adv-KIP} CIFAR-10 \textbf{TRADES} dataset from the \texttt{FC3} kernel, with 40,000 samples. Setting: $\ell_\infty$, $\epsilon=8/255.$}
    \label{tab:nn-modern-TRADES-40K}
\end{table}

\begin{figure}
    \centering
    \includegraphics[scale=0.4]{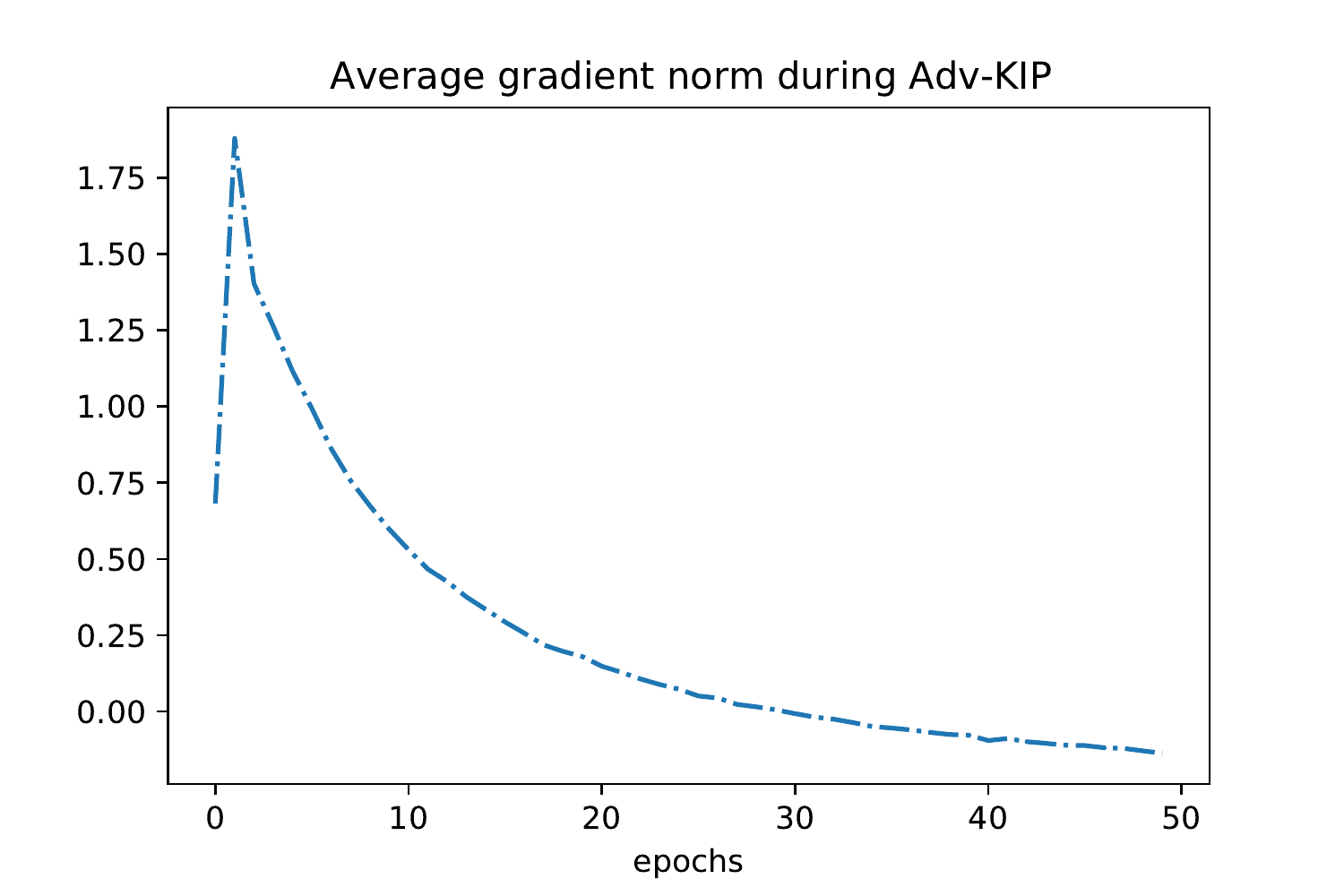}
    \caption{The average gradient norm of an FC3 kernel on a validation set during the distillation procedure of Algorithm \ref{alg:adv-kip}. We see that the training data evolves to cause gradient shrinkage of the model. Setting: CIFAR-10, FC3, $|\mathcal{X}_S|$ = 40k, $|\mathcal{X}_T|$ = 10k, 10 PGD steps, cross entropy loss in outer loop.}
    \label{fig:avg_grad_advkip}
\end{figure}

\begin{figure}
    \centering
    \includegraphics[scale=0.32]{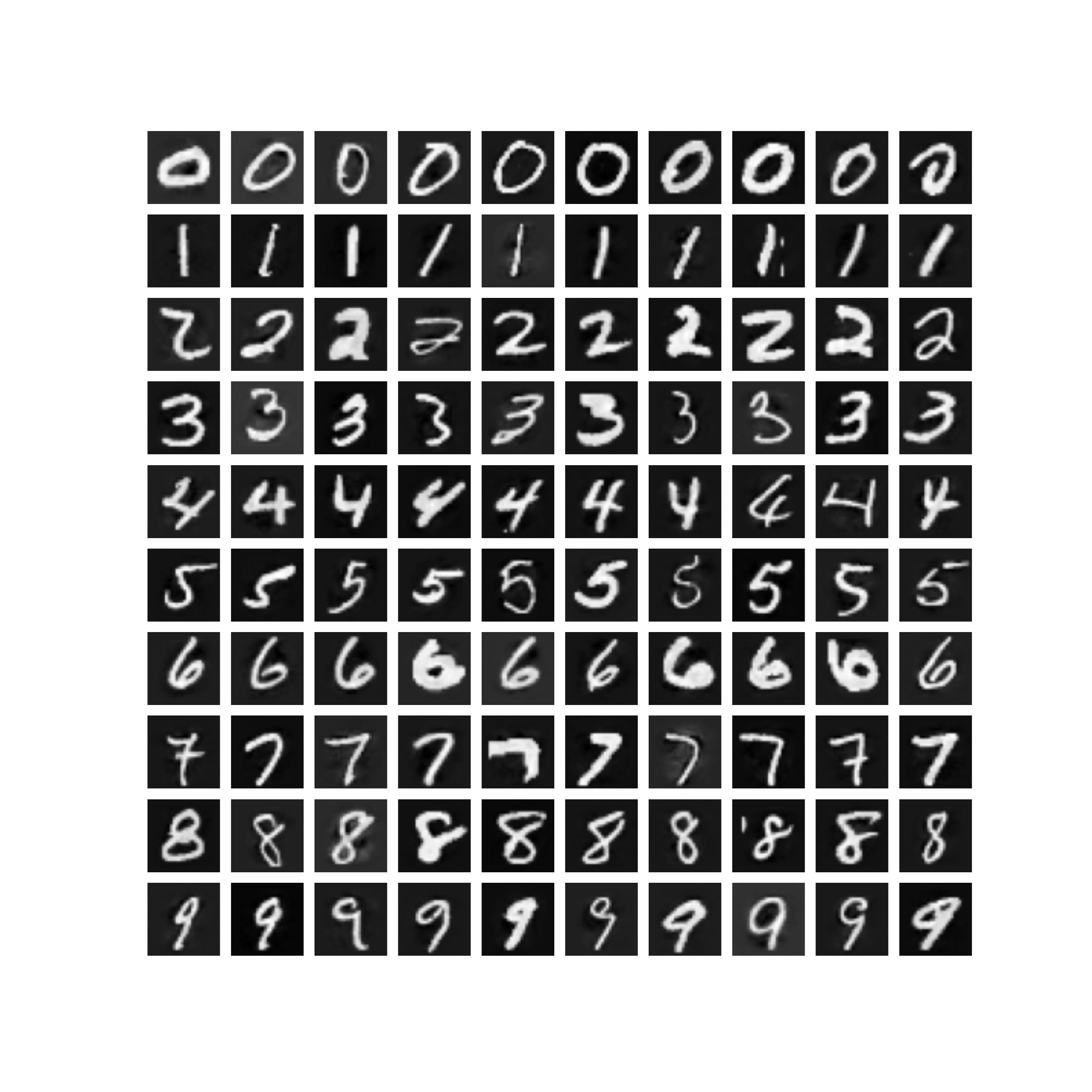}
    \includegraphics[scale=0.32]{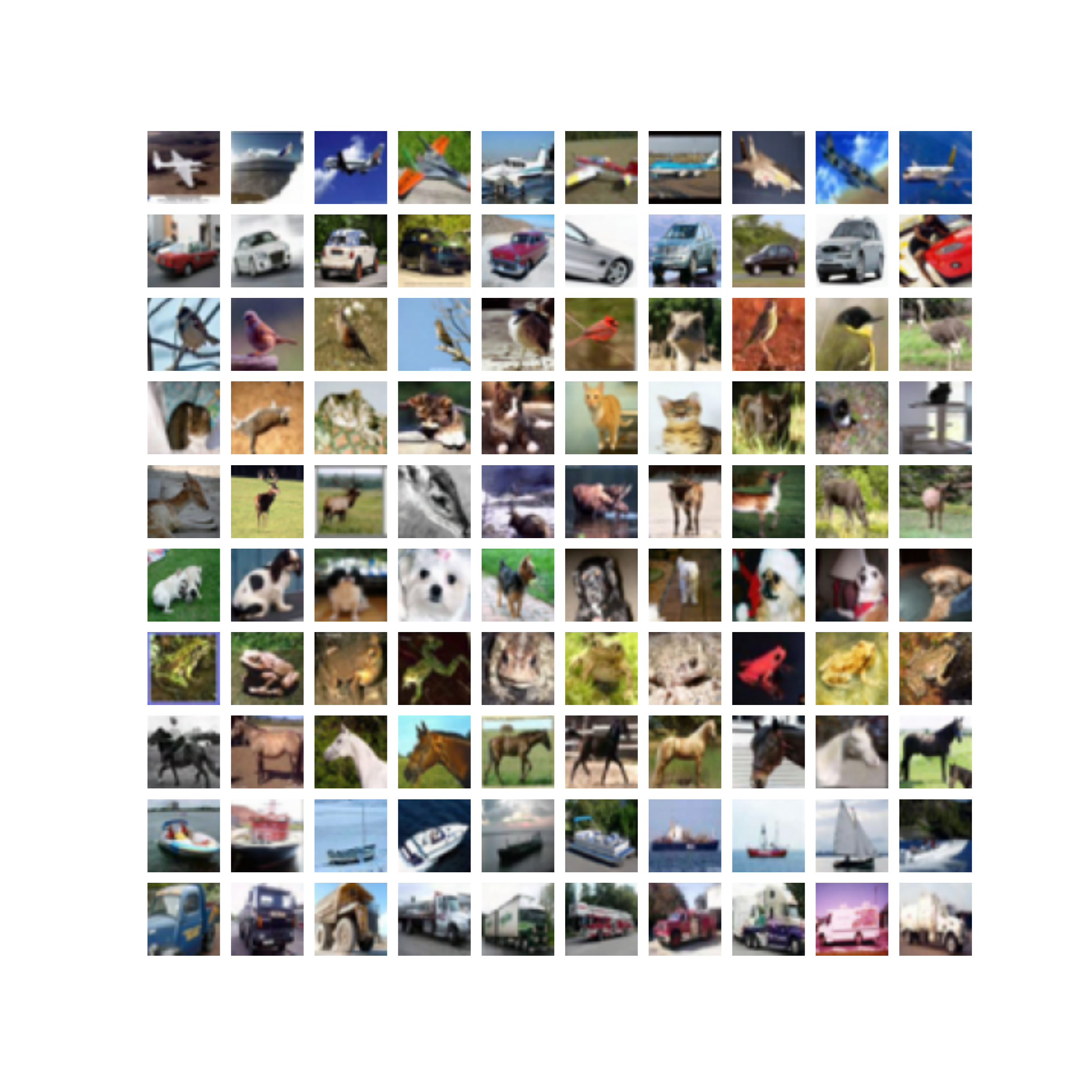}
    \caption{MNIST (left) and CIFAR-10 (right) distilled images with trained labels from an \texttt{FC7} kernel.}
    \label{fig:mnist_images}
\end{figure}

\end{document}

%% file: math_commands.tex
\usepackage{amsmath,amsfonts,bm}

\def\eqref#1{equation~\ref{#1}}

\def\1{\bm{1}}

\DeclareMathAlphabet{\mathsfit}{\encodingdefault}{\sfdefault}{m}{sl}
\SetMathAlphabet{\mathsfit}{bold}{\encodingdefault}{\sfdefault}{bx}{n}













%% file: icml2023.bbl
\begin{thebibliography}{55}
\providecommand{\natexlab}[1]{#1}
\providecommand{\url}[1]{\texttt{#1}}
\expandafter\ifx\csname urlstyle\endcsname\relax
  \providecommand{\doi}[1]{doi: #1}\else
  \providecommand{\doi}{doi: \begingroup \urlstyle{rm}\Url}\fi

\bibitem[Allen-Zhu \& Li(2022)Allen-Zhu and Li]{allen2022feature}
Allen-Zhu, Z. and Li, Y.
\newblock Feature purification: How adversarial training performs robust deep
  learning.
\newblock In \emph{2021 IEEE 62nd Annual Symposium on Foundations of Computer
  Science (FOCS)}, pp.\  977--988. IEEE, 2022.

\bibitem[Arora et~al.(2019)Arora, Du, Hu, Li, Salakhutdinov, and Wang]{Aro+19b}
Arora, S., Du, S.~S., Hu, W., Li, Z., Salakhutdinov, R., and Wang, R.
\newblock On exact computation with an infinitely wide neural net.
\newblock In Wallach, H.~M., Larochelle, H., Beygelzimer, A.,
  d'Alch{\'{e}}{-}Buc, F., Fox, E.~B., and Garnett, R. (eds.), \emph{Advances
  in Neural Information Processing Systems 32: Annual Conference on Neural
  Information Processing Systems 2019, NeurIPS 2019, December 8-14, 2019,
  Vancouver, BC, Canada}, pp.\  8139--8148, 2019.

\bibitem[Athalye et~al.(2018)Athalye, Carlini, and Wagner]{Ath+18}
Athalye, A., Carlini, N., and Wagner, D.~A.
\newblock Obfuscated gradients give a false sense of security: Circumventing
  defenses to adversarial examples.
\newblock In Dy, J.~G. and Krause, A. (eds.), \emph{Proceedings of the 35th
  International Conference on Machine Learning, {ICML} 2018,
  Stockholmsm{\"{a}}ssan, Stockholm, Sweden, July 10-15, 2018}, volume~80 of
  \emph{Proceedings of Machine Learning Research}, pp.\  274--283. {PMLR},
  2018.

\bibitem[Awasthi et~al.(2021)Awasthi, Chatziafratis, Chen, and
  Vijayaraghavan]{Awa+21}
Awasthi, P., Chatziafratis, V., Chen, X., and Vijayaraghavan, A.
\newblock Adversarially robust low dimensional representations.
\newblock In Belkin, M. and Kpotufe, S. (eds.), \emph{Conference on Learning
  Theory, {COLT} 2021, 15-19 August 2021, Boulder, Colorado, {USA}}, volume 134
  of \emph{Proceedings of Machine Learning Research}, pp.\  237--325. {PMLR},
  2021.

\bibitem[Bradbury et~al.(2018)Bradbury, Frostig, Hawkins, Johnson, Leary,
  Maclaurin, Necula, Paszke, Vander{P}las, Wanderman-{M}ilne, and
  Zhang]{Bra+18}
Bradbury, J., Frostig, R., Hawkins, P., Johnson, M.~J., Leary, C., Maclaurin,
  D., Necula, G., Paszke, A., Vander{P}las, J., Wanderman-{M}ilne, S., and
  Zhang, Q.
\newblock {JAX}: composable transformations of {P}ython+{N}um{P}y programs,
  2018.
\newblock URL \url{http://github.com/google/jax}.

\bibitem[Carlini \& Wagner(2017)Carlini and Wagner]{CaWa17}
Carlini, N. and Wagner, D.~A.
\newblock Towards evaluating the robustness of neural networks.
\newblock In \emph{2017 {IEEE} Symposium on Security and Privacy, {SP} 2017,
  San Jose, CA, USA, May 22-26, 2017}, pp.\  39--57. {IEEE} Computer Society,
  2017.

\bibitem[Chen et~al.(2021)Chen, Gong, and Wang]{Chen+21}
Chen, W., Gong, X., and Wang, Z.
\newblock Neural architecture search on imagenet in four {GPU} hours: {A}
  theoretically inspired perspective.
\newblock In \emph{9th International Conference on Learning Representations,
  {ICLR} 2021, Virtual Event, Austria, May 3-7, 2021}. OpenReview.net, 2021.

\bibitem[Croce \& Hein(2020)Croce and Hein]{croce2020reliable}
Croce, F. and Hein, M.
\newblock Reliable evaluation of adversarial robustness with an ensemble of
  diverse parameter-free attacks.
\newblock In \emph{ICML}, 2020.

\bibitem[Croce \& Hein(2021)Croce and Hein]{croce2021mind}
Croce, F. and Hein, M.
\newblock Mind the box: $l_1$-apgd for sparse adversarial attacks on image
  classifiers.
\newblock In \emph{ICML}, 2021.

\bibitem[Croce et~al.(2020)Croce, Andriushchenko, Sehwag, Debenedetti,
  Flammarion, Chiang, Mittal, and Hein]{Cro+20}
Croce, F., Andriushchenko, M., Sehwag, V., Debenedetti, E., Flammarion, N.,
  Chiang, M., Mittal, P., and Hein, M.
\newblock Robustbench: a standardized adversarial robustness benchmark.
\newblock \emph{arXiv preprint arXiv:2010.09670}, 2020.

\bibitem[DeGroot \& Fienberg(1983)DeGroot and Fienberg]{degroot1983comparison}
DeGroot, M.~H. and Fienberg, S.~E.
\newblock The comparison and evaluation of forecasters.
\newblock \emph{Journal of the Royal Statistical Society: Series D (The
  Statistician)}, 32\penalty0 (1-2):\penalty0 12--22, 1983.

\bibitem[Deng(2012)]{deng2012mnist}
Deng, L.
\newblock The mnist database of handwritten digit images for machine learning
  research.
\newblock \emph{IEEE Signal Processing Magazine}, 29\penalty0 (6):\penalty0
  141--142, 2012.

\bibitem[Garg et~al.(2018)Garg, Sharan, Zhang, and Valiant]{Gar+18}
Garg, S., Sharan, V., Zhang, B.~H., and Valiant, G.
\newblock A spectral view of adversarially robust features.
\newblock In Bengio, S., Wallach, H.~M., Larochelle, H., Grauman, K.,
  Cesa{-}Bianchi, N., and Garnett, R. (eds.), \emph{Advances in Neural
  Information Processing Systems 31: Annual Conference on Neural Information
  Processing Systems 2018, NeurIPS 2018, December 3-8, 2018, Montr{\'{e}}al,
  Canada}, pp.\  10159--10169, 2018.

\bibitem[Goodfellow et~al.(2015)Goodfellow, Shlens, and Szegedy]{GSS15}
Goodfellow, I.~J., Shlens, J., and Szegedy, C.
\newblock Explaining and harnessing adversarial examples.
\newblock In Bengio, Y. and LeCun, Y. (eds.), \emph{3rd International
  Conference on Learning Representations, {ICLR} 2015, San Diego, CA, USA, May
  7-9, 2015, Conference Track Proceedings}, 2015.

\bibitem[Guo et~al.(2017)Guo, Pleiss, Sun, and Weinberger]{guo2017calibration}
Guo, C., Pleiss, G., Sun, Y., and Weinberger, K.~Q.
\newblock On calibration of modern neural networks.
\newblock In \emph{International conference on machine learning}, pp.\
  1321--1330. PMLR, 2017.

\bibitem[Hashimoto et~al.(2018)Hashimoto, Srivastava, Namkoong, and
  Liang]{Hash+18}
Hashimoto, T.~B., Srivastava, M., Namkoong, H., and Liang, P.
\newblock Fairness without demographics in repeated loss minimization.
\newblock In Dy, J.~G. and Krause, A. (eds.), \emph{Proceedings of the 35th
  International Conference on Machine Learning, {ICML} 2018,
  Stockholmsm{\"{a}}ssan, Stockholm, Sweden, July 10-15, 2018}, Proceedings of
  Machine Learning Research, 2018.

\bibitem[Havasi et~al.(2020)Havasi, Jenatton, Fort, Liu, Snoek,
  Lakshminarayanan, Dai, and Tran]{havasi2020training}
Havasi, M., Jenatton, R., Fort, S., Liu, J.~Z., Snoek, J., Lakshminarayanan,
  B., Dai, A.~M., and Tran, D.
\newblock Training independent subnetworks for robust prediction.
\newblock \emph{arXiv preprint arXiv:2010.06610}, 2020.

\bibitem[He et~al.(2016)He, Zhang, Ren, and Sun]{he2016deep}
He, K., Zhang, X., Ren, S., and Sun, J.
\newblock Deep residual learning for image recognition.
\newblock In \emph{Proceedings of the IEEE conference on computer vision and
  pattern recognition}, pp.\  770--778, 2016.

\bibitem[Hendrycks \& Dietterich(2019)Hendrycks and Dietterich]{HeDi19}
Hendrycks, D. and Dietterich, T.~G.
\newblock Benchmarking neural network robustness to common corruptions and
  perturbations.
\newblock In \emph{7th International Conference on Learning Representations,
  {ICLR} 2019, New Orleans, LA, USA, May 6-9, 2019}, 2019.

\bibitem[Ilyas et~al.(2019)Ilyas, Santurkar, Tsipras, Engstrom, Tran, and
  Madry]{Ily+19}
Ilyas, A., Santurkar, S., Tsipras, D., Engstrom, L., Tran, B., and Madry, A.
\newblock Adversarial examples are not bugs, they are features.
\newblock In Wallach, H.~M., Larochelle, H., Beygelzimer, A.,
  d'Alch{\'{e}}{-}Buc, F., Fox, E.~B., and Garnett, R. (eds.), \emph{Advances
  in Neural Information Processing Systems 32: Annual Conference on Neural
  Information Processing Systems 2019, NeurIPS 2019, December 8-14, 2019,
  Vancouver, BC, Canada}, pp.\  125--136, 2019.

\bibitem[Jacot et~al.(2018)Jacot, Hongler, and Gabriel]{JHG18}
Jacot, A., Hongler, C., and Gabriel, F.
\newblock Neural tangent kernel: Convergence and generalization in neural
  networks.
\newblock In Bengio, S., Wallach, H.~M., Larochelle, H., Grauman, K.,
  Cesa{-}Bianchi, N., and Garnett, R. (eds.), \emph{Advances in Neural
  Information Processing Systems 31: Annual Conference on Neural Information
  Processing Systems 2018, NeurIPS 2018, December 3-8, 2018, Montr{\'{e}}al,
  Canada}, pp.\  8580--8589, 2018.

\bibitem[Kingma \& Ba(2015)Kingma and Ba]{KiBa15}
Kingma, D.~P. and Ba, J.
\newblock Adam: {A} method for stochastic optimization.
\newblock In Bengio, Y. and LeCun, Y. (eds.), \emph{3rd International
  Conference on Learning Representations, {ICLR} 2015, San Diego, CA, USA, May
  7-9, 2015, Conference Track Proceedings}, 2015.

\bibitem[Krizhevsky(2009)]{Krizhevsky09CIFAR}
Krizhevsky, A.
\newblock Learning multiple layers of features from tiny images.
\newblock Technical report, 2009.

\bibitem[Krizhevsky et~al.(2012)Krizhevsky, Sutskever, and Hinton]{AlexNet}
Krizhevsky, A., Sutskever, I., and Hinton, G.~E.
\newblock Imagenet classification with deep convolutional neural networks.
\newblock In Pereira, F., Burges, C., Bottou, L., and Weinberger, K. (eds.),
  \emph{Advances in Neural Information Processing Systems}, volume~25. Curran
  Associates, Inc., 2012.

\bibitem[Kurakin et~al.(2017)Kurakin, Goodfellow, and Bengio]{KGB17}
Kurakin, A., Goodfellow, I.~J., and Bengio, S.
\newblock Adversarial examples in the physical world.
\newblock In \emph{5th International Conference on Learning Representations,
  {ICLR} 2017, Toulon, France, April 24-26, 2017, Workshop Track Proceedings},
  2017.

\bibitem[Lakshminarayanan et~al.(2017)Lakshminarayanan, Pritzel, and
  Blundell]{lakshminarayanan2017simple}
Lakshminarayanan, B., Pritzel, A., and Blundell, C.
\newblock Simple and scalable predictive uncertainty estimation using deep
  ensembles.
\newblock \emph{Advances in neural information processing systems}, 30, 2017.

\bibitem[Lee et~al.(2019)Lee, Xiao, Schoenholz, Bahri, Novak, Sohl-Dickstein,
  and Pennington]{Lee+19}
Lee, J., Xiao, L., Schoenholz, S., Bahri, Y., Novak, R., Sohl-Dickstein, J.,
  and Pennington, J.
\newblock {Wide Neural Networks of Any Depth Evolve as Linear Models Under
  Gradient Descent}.
\newblock In Wallach, H., Larochelle, H., Beygelzimer, A., d\textquotesingle
  Alch\'{e}-Buc, F., Fox, E., and Garnett, R. (eds.), \emph{Advances in Neural
  Information Processing Systems}, volume~32. Curran Associates, Inc., 2019.

\bibitem[Madry et~al.(2018)Madry, Makelov, Schmidt, Tsipras, and Vladu]{Mad+18}
Madry, A., Makelov, A., Schmidt, L., Tsipras, D., and Vladu, A.
\newblock {Towards Deep Learning Models Resistant to Adversarial Attacks}.
\newblock In \emph{International Conference on Learning Representations}, 2018.

\bibitem[Moosavi{-}Dezfooli et~al.(2017)Moosavi{-}Dezfooli, Fawzi, Fawzi, and
  Frossard]{Dez+17}
Moosavi{-}Dezfooli, S., Fawzi, A., Fawzi, O., and Frossard, P.
\newblock Universal adversarial perturbations.
\newblock In \emph{2017 {IEEE} Conference on Computer Vision and Pattern
  Recognition, {CVPR} 2017, Honolulu, HI, USA, July 21-26, 2017}, pp.\  86--94.
  {IEEE} Computer Society, 2017.

\bibitem[Naeini et~al.(2015)Naeini, Cooper, and
  Hauskrecht]{naeini2015obtaining}
Naeini, M.~P., Cooper, G., and Hauskrecht, M.
\newblock Obtaining well calibrated probabilities using bayesian binning.
\newblock In \emph{Twenty-Ninth AAAI Conference on Artificial Intelligence},
  2015.

\bibitem[Nguyen et~al.(2021{\natexlab{a}})Nguyen, Chen, and Lee]{Ngu+21a}
Nguyen, T., Chen, Z., and Lee, J.
\newblock Dataset meta-learning from kernel ridge-regression.
\newblock In \emph{9th International Conference on Learning Representations,
  {ICLR} 2021, Virtual Event, Austria, May 3-7, 2021}. OpenReview.net,
  2021{\natexlab{a}}.

\bibitem[Nguyen et~al.(2021{\natexlab{b}})Nguyen, Novak, Xiao, and
  Lee]{Ngu+21b}
Nguyen, T., Novak, R., Xiao, L., and Lee, J.
\newblock Dataset distillation with infinitely wide convolutional networks.
\newblock \emph{CoRR}, abs/2107.13034, 2021{\natexlab{b}}.

\bibitem[Niculescu-Mizil \& Caruana(2005)Niculescu-Mizil and
  Caruana]{niculescu2005predicting}
Niculescu-Mizil, A. and Caruana, R.
\newblock Predicting good probabilities with supervised learning.
\newblock In \emph{Proceedings of the 22nd international conference on Machine
  learning}, pp.\  625--632, 2005.

\bibitem[Novak et~al.(2020)Novak, Xiao, Hron, Lee, Alemi, Sohl{-}Dickstein, and
  Schoenholz]{Nov+20}
Novak, R., Xiao, L., Hron, J., Lee, J., Alemi, A.~A., Sohl{-}Dickstein, J., and
  Schoenholz, S.~S.
\newblock Neural tangents: Fast and easy infinite neural networks in python.
\newblock In \emph{8th International Conference on Learning Representations,
  {ICLR} 2020, Addis Ababa, Ethiopia, April 26-30, 2020}. OpenReview.net, 2020.

\bibitem[Papernot et~al.(2016)Papernot, McDaniel, Wu, Jha, and Swami]{Pap+16}
Papernot, N., McDaniel, P.~D., Wu, X., Jha, S., and Swami, A.
\newblock Distillation as a defense to adversarial perturbations against deep
  neural networks.
\newblock In \emph{{IEEE} Symposium on Security and Privacy, {SP} 2016, San
  Jose, CA, USA, May 22-26, 2016}, pp.\  582--597. {IEEE} Computer Society,
  2016.
\newblock \doi{10.1109/SP.2016.41}.
\newblock URL \url{https://doi.org/10.1109/SP.2016.41}.

\bibitem[Papernot et~al.(2017)Papernot, McDaniel, Goodfellow, Jha, Celik, and
  Swami]{Pap+17}
Papernot, N., McDaniel, P.~D., Goodfellow, I.~J., Jha, S., Celik, Z.~B., and
  Swami, A.
\newblock Practical black-box attacks against machine learning.
\newblock In Karri, R., Sinanoglu, O., Sadeghi, A., and Yi, X. (eds.),
  \emph{Proceedings of the 2017 {ACM} on Asia Conference on Computer and
  Communications Security, AsiaCCS 2017, Abu Dhabi, United Arab Emirates, April
  2-6, 2017}, pp.\  506--519. {ACM}, 2017.

\bibitem[Papernot et~al.(2018)Papernot, Faghri, Carlini, Goodfellow, Feinman,
  Kurakin, Xie, Sharma, Brown, Roy, Matyasko, Behzadan, Hambardzumyan, Zhang,
  Juang, Li, Sheatsley, Garg, Uesato, Gierke, Dong, Berthelot, Hendricks,
  Rauber, and Long]{Pap+18}
Papernot, N., Faghri, F., Carlini, N., Goodfellow, I., Feinman, R., Kurakin,
  A., Xie, C., Sharma, Y., Brown, T., Roy, A., Matyasko, A., Behzadan, V.,
  Hambardzumyan, K., Zhang, Z., Juang, Y.-L., Li, Z., Sheatsley, R., Garg, A.,
  Uesato, J., Gierke, W., Dong, Y., Berthelot, D., Hendricks, P., Rauber, J.,
  and Long, R.
\newblock Technical report on the cleverhans v2.1.0 adversarial examples
  library.
\newblock \emph{arXiv preprint arXiv:1610.00768}, 2018.

\bibitem[Paul et~al.(2021)Paul, Ganguli, and Dziugaite]{Diet21}
Paul, M., Ganguli, S., and Dziugaite, G.~K.
\newblock Deep learning on a data diet: Finding important examples early in
  training.
\newblock In Ranzato, M., Beygelzimer, A., Dauphin, Y., Liang, P., and Vaughan,
  J.~W. (eds.), \emph{Advances in Neural Information Processing Systems},
  volume~34, pp.\  20596--20607. Curran Associates, Inc., 2021.

\bibitem[Paul et~al.(2022)Paul, Larsen, Ganguli, Frankle, and
  Dziugaite]{paul2022lottery}
Paul, M., Larsen, B.~W., Ganguli, S., Frankle, J., and Dziugaite, G.~K.
\newblock Lottery tickets on a data diet: Finding initializations with sparse
  trainable networks.
\newblock In Oh, A.~H., Agarwal, A., Belgrave, D., and Cho, K. (eds.),
  \emph{Advances in Neural Information Processing Systems}, 2022.
\newblock URL \url{https://openreview.net/forum?id=QLPzCpu756J}.

\bibitem[Shafahi et~al.(2019)Shafahi, Najibi, Ghiasi, Xu, Dickerson, Studer,
  Davis, Taylor, and Goldstein]{Sha+19}
Shafahi, A., Najibi, M., Ghiasi, A., Xu, Z., Dickerson, J.~P., Studer, C.,
  Davis, L.~S., Taylor, G., and Goldstein, T.
\newblock Adversarial training for free!
\newblock In Wallach, H.~M., Larochelle, H., Beygelzimer, A.,
  d'Alch{\'{e}}{-}Buc, F., Fox, E.~B., and Garnett, R. (eds.), \emph{Advances
  in Neural Information Processing Systems 32: Annual Conference on Neural
  Information Processing Systems 2019, NeurIPS 2019, December 8-14, 2019,
  Vancouver, BC, Canada}, pp.\  3353--3364, 2019.

\bibitem[Simonyan \& Zisserman(2015)Simonyan and Zisserman]{Simonyan15}
Simonyan, K. and Zisserman, A.
\newblock Very deep convolutional networks for large-scale image recognition.
\newblock In \emph{International Conference on Learning Representations}, 2015.

\bibitem[Sinha et~al.(2018)Sinha, Namkoong, and Duchi]{Sin+18}
Sinha, A., Namkoong, H., and Duchi, J.~C.
\newblock Certifying some distributional robustness with principled adversarial
  training.
\newblock In \emph{6th International Conference on Learning Representations,
  {ICLR} 2018, Vancouver, BC, Canada, April 30 - May 3, 2018, Conference Track
  Proceedings}. OpenReview.net, 2018.

\bibitem[Sorscher et~al.(2022)Sorscher, Geirhos, Shekhar, Ganguli, and
  Morcos]{sorscher2022beyond}
Sorscher, B., Geirhos, R., Shekhar, S., Ganguli, S., and Morcos, A.~S.
\newblock Beyond neural scaling laws: beating power law scaling via data
  pruning.
\newblock In Oh, A.~H., Agarwal, A., Belgrave, D., and Cho, K. (eds.),
  \emph{Advances in Neural Information Processing Systems}, 2022.
\newblock URL \url{https://openreview.net/forum?id=UmvSlP-PyV}.

\bibitem[Szegedy et~al.(2014)Szegedy, Zaremba, Sutskever, Bruna, Erhan,
  Goodfellow, and Fergus]{Sze+14}
Szegedy, C., Zaremba, W., Sutskever, I., Bruna, J., Erhan, D., Goodfellow, I.,
  and Fergus, R.
\newblock Intriguing properties of neural networks, 2014.

\bibitem[Tancik et~al.(2020)Tancik, Srinivasan, Mildenhall, Fridovich{-}Keil,
  Raghavan, Singhal, Ramamoorthi, Barron, and Ng]{Tan+20}
Tancik, M., Srinivasan, P.~P., Mildenhall, B., Fridovich{-}Keil, S., Raghavan,
  N., Singhal, U., Ramamoorthi, R., Barron, J.~T., and Ng, R.
\newblock Fourier features let networks learn high frequency functions in low
  dimensional domains.
\newblock In Larochelle, H., Ranzato, M., Hadsell, R., Balcan, M., and Lin, H.
  (eds.), \emph{Advances in Neural Information Processing Systems 33: Annual
  Conference on Neural Information Processing Systems 2020, NeurIPS 2020,
  December 6-12, 2020, virtual}, 2020.

\bibitem[Tsilivis \& Kempe(2022)Tsilivis and Kempe]{TsilivisKempe22}
Tsilivis, N. and Kempe, J.
\newblock What can the neural tangent kernel tell us about adversarial
  robustness?
\newblock In \emph{Advances in Neural Information Processing Systems}. Curran
  Associates, Inc., 2022.

\bibitem[Tsipras et~al.(2019)Tsipras, Santurkar, Engstrom, Turner, and
  Madry]{Tsi+19}
Tsipras, D., Santurkar, S., Engstrom, L., Turner, A., and Madry, A.
\newblock Robustness may be at odds with accuracy.
\newblock In \emph{7th International Conference on Learning Representations,
  {ICLR} 2019, New Orleans, LA, USA, May 6-9, 2019}, 2019.

\bibitem[Volpi et~al.(2018)Volpi, Namkoong, Sener, Duchi, Murino, and
  Savarese]{Vol+18}
Volpi, R., Namkoong, H., Sener, O., Duchi, J.~C., Murino, V., and Savarese, S.
\newblock Generalizing to unseen domains via adversarial data augmentation.
\newblock In Bengio, S., Wallach, H.~M., Larochelle, H., Grauman, K.,
  Cesa{-}Bianchi, N., and Garnett, R. (eds.), \emph{Advances in Neural
  Information Processing Systems 31: Annual Conference on Neural Information
  Processing Systems 2018, NeurIPS 2018, December 3-8, 2018, Montr{\'{e}}al,
  Canada}, pp.\  5339--5349, 2018.

\bibitem[Wang et~al.(2018{\natexlab{a}})Wang, Zhu, Torralba, and Efros]{Wan+18}
Wang, T., Zhu, J., Torralba, A., and Efros, A.~A.
\newblock Dataset distillation.
\newblock \emph{CoRR}, abs/1811.10959, 2018{\natexlab{a}}.

\bibitem[Wang et~al.(2018{\natexlab{b}})Wang, Zhu, Torralba, and Efros]{Wang18}
Wang, T., Zhu, J.-Y., Torralba, A., and Efros, A.~A.
\newblock Dataset distillation, 2018{\natexlab{b}}.
\newblock URL \url{https://arxiv.org/abs/1811.10959}.

\bibitem[Wenzel et~al.(2020)Wenzel, Snoek, Tran, and
  Jenatton]{wenzel2020hyperparameter}
Wenzel, F., Snoek, J., Tran, D., and Jenatton, R.
\newblock Hyperparameter ensembles for robustness and uncertainty
  quantification.
\newblock \emph{Advances in Neural Information Processing Systems},
  33:\penalty0 6514--6527, 2020.

\bibitem[Wong et~al.(2020{\natexlab{a}})Wong, Rice, and Kolter]{WRK20}
Wong, E., Rice, L., and Kolter, J.~Z.
\newblock Fast is better than free: Revisiting adversarial training.
\newblock In \emph{8th International Conference on Learning Representations,
  {ICLR} 2020, Addis Ababa, Ethiopia, April 26-30, 2020}, 2020{\natexlab{a}}.

\bibitem[Wong et~al.(2020{\natexlab{b}})Wong, Rice, and Kolter]{Wong+20}
Wong, E., Rice, L., and Kolter, J.~Z.
\newblock Fast is better than free: Revisiting adversarial training.
\newblock In \emph{8th International Conference on Learning Representations,
  {ICLR} 2020, Addis Ababa, Ethiopia, April 26-30, 2020}. OpenReview.net,
  2020{\natexlab{b}}.

\bibitem[Yuan \& Wu(2021)Yuan and Wu]{YuWu21}
Yuan, C.-H. and Wu, S.-H.
\newblock Neural tangent generalization attacks.
\newblock In Meila, M. and Zhang, T. (eds.), \emph{Proceedings of the 38th
  International Conference on Machine Learning}, volume 139 of
  \emph{Proceedings of Machine Learning Research}, pp.\  12230--12240. PMLR,
  18--24 Jul 2021.

\bibitem[Zhang et~al.(2019)Zhang, Yu, Jiao, Xing, Ghaoui, and Jordan]{Zha+19}
Zhang, H., Yu, Y., Jiao, J., Xing, E.~P., Ghaoui, L.~E., and Jordan, M.~I.
\newblock Theoretically principled trade-off between robustness and accuracy.
\newblock In Chaudhuri, K. and Salakhutdinov, R. (eds.), \emph{Proceedings of
  the 36th International Conference on Machine Learning, {ICML} 2019, 9-15 June
  2019, Long Beach, California, {USA}}, volume~97 of \emph{Proceedings of
  Machine Learning Research}, pp.\  7472--7482. {PMLR}, 2019.

\end{thebibliography}
